\documentclass{article}

\PassOptionsToPackage{numbers, compress}{natbib}

\usepackage[preprint]{neurips_2021}

\usepackage[utf8]{inputenc} 
\usepackage[T1]{fontenc}    
\usepackage{hyperref}       
\usepackage{url}            
\usepackage{booktabs}       
\usepackage{amsfonts}       
\usepackage{nicefrac}       
\usepackage{microtype}      
\usepackage{xcolor}         

\usepackage{multirow}
\usepackage{multicol}
\usepackage{dsfont}
\usepackage{colortbl}
\usepackage{lipsum}
\usepackage{amsthm}
\usepackage{mathrsfs}
\usepackage{amsmath}
\usepackage{bm}
\usepackage{algorithm}
\usepackage{algorithmicx}
\usepackage{algorithmicx}
\usepackage{algorithm} 
\usepackage[noend]{algpseudocode} 
\usepackage{graphicx}
\usepackage{wrapfig}
\usepackage{picinpar}
\usepackage{makecell}
\usepackage{enumitem}
\setitemize{noitemsep,topsep=0pt,parsep=0pt,partopsep=0pt}

\newtheorem{fact}{Fact}
\newtheorem{observation}{Observation}
\newtheorem{definition}{Definition}

\newtheorem{property}{Property}
\newcommand{\vpara}[1]{\vspace{0.07in}\noindent\textbf{#1}}
\newcommand{\todo}[1]{\textcolor{red}{todo: #1}}

\newcommand{\tabincell}[2]{\begin{tabular}{@{}#1@{}}#2\end{tabular}}

\newcommand{\HGCPAM}{\text{HGC\_AdaM}}
\newcommand{\HGC}{\text{HGC}}
\newcommand{\AM}{\text{AdaM}}
\newcommand{\GCC}{\text{GCC}}

\newcommand{\hgc}{\text{HGC}}
\newcommand{\am}{\text{AdaM}}
\newcommand{\rdm}{\text{RdM}}
\newcommand{\hgcn}{\text{HGC}}
\newcommand{\amn}{\text{AdaM}}

\newcommand{\SSAD}{\text{AP\_NF}}
\newcommand{\BSGD}{\text{SocS\_NF}}
\newcommand{\SGCD}{\text{SocL\_NF}}
\newcommand{\MolD}{\text{MolD}}
\newcommand{\FOS}{\text{FO}}
\newcommand{\HOS}{\text{HO}}

\newcommand{\eat}[1]{}
\def\ie{\emph{i.e}\onedot}
\definecolor{mygray}{gray}{.9}
\definecolor{mypink}{rgb}{.99,.91,.95}
\definecolor{mycyan}{cmyk}{.3,0,0,0}

\title{Similarity-aware Positive Instance Sampling for Graph Contrastive Pre-training}

\author{%
  Xueyi Liu$^{1}$, Yu Rong$^{2}$, Tingyang Xu$^{2}$, Fuchun Sun$^{1}$, Wenbing Huang$^{1}$, Junzhou Huang$^{2}$ \\
  $^{1}$Department of Computer Science and Technology, Tsinghua University \\ $^{2}$Tencent AI Lab
}

\begin{document}

\maketitle

\begin{abstract}
  Graph instance contrastive learning has been proved as an effective task for Graph Neural Network (GNN) pre-training. 
  However, one key issue may seriously impede the representative power in existing works: Positive instances created by current methods often miss  crucial information of graphs or even yield illegal instances (such as non-chemically-aware graphs in molecular generation). To remedy this issue,  we propose to select positive graph 
 instances directly from existing graphs in the training set, which ultimately maintains the legality and similarity to the target graphs. Our selection is based on certain domain-specific pair-wise similarity measurements as well as sampling from a hierarchical graph encoding similarity relations among graphs. 
 Besides, we develop an adaptive node-level pre-training method to dynamically mask nodes to distribute them evenly in the graph. 
We conduct extensive experiments on $13$ graph classification and node classification benchmark datasets from various domains. The results demonstrate that the GNN models pre-trained by our strategies 
can outperform those trained-from-scratch models as well as the variants obtained by existing methods. 
\end{abstract}

\section{Introduction} \label{sec_intro}

Pre-training on graph data has received wide interests in recent years, with a large range of insightful works focused on learning universal graph structural patterns lying in different kinds of graph data~\citep{qiu2020gcc,hu2019strategies,you2020graph,rong2020grover}. For instance, Hu et al.~\citep{hu2019strategies} pre-train graph neural networks on molecules and transfer the learned model to molecular graph classification tasks, while Qiu et al.~\citep{qiu2020gcc} pioneer  pre-training on big graphs. 
Compared with traditional 
semi-supervised or supervised training methods for graph neural networks~\citep{hamilton2017inductive,kipf2016semi,xu2018powerful,gilmer2017neural,velivckovic2017graph}, pre-training tasks formulate the training objective without the access of training labels, and they empower graph neural networks to be generalized to unseen graphs or nodes with no or minor fine-tuning training cost. How to define proper pre-training tasks comes as the principal and also the most challenging part in graph self-supervised learning. 

Among current works, graph instance contrastive learning based pre-training tasks have been proved effective to learn graph strcutrual information~\citep{qiu2020gcc,you2020graph}.
It preforms contrast between positive/negative instance pairs extracted from real graphs observed in the dataset. 
Though positive pairs for graph contrastive learning seems easy to define for those tasks not performed on graph instances, like DeepWalk~\citep{perozzi2014deepwalk}, node2vec~\citep{grover2016node2vec}, where near node-node pairs are treated as positive pairs and Infomax based models like DGI~\citep{velickovic2019deep}, InfoGraph~\citep{sun2019infograph}, where node-graph pairs from a same graph are treated as positive pairs, it is not the case for graph instance contrastive learning.
Attempts from previous literature mainly focus on devising suitable graph augmentation methods, such as graph sampling~\citep{qiu2020gcc,you2020graph}, node dropping~\citep{you2020graph}, edge perturbation~\citep{you2020graph}, and diffusion graph~\citep{hassani2020contrastive} to get positive graph instances from the original graph. 
Despite the achievements they have made using such graph data augmentation strategies, 
we assume that such perturbation based graph data augmentation methods are not universal strategies to get ideal positive samples preserving \emph{necessary information} for graph contrastive learning for various kinds of graph data such as molecular graphs, social graphs, and academic graphs.

\begin{wrapfigure}[16]{R}{0.48\textwidth}
  \vspace{-4ex}
\includegraphics[width=0.48\textwidth]{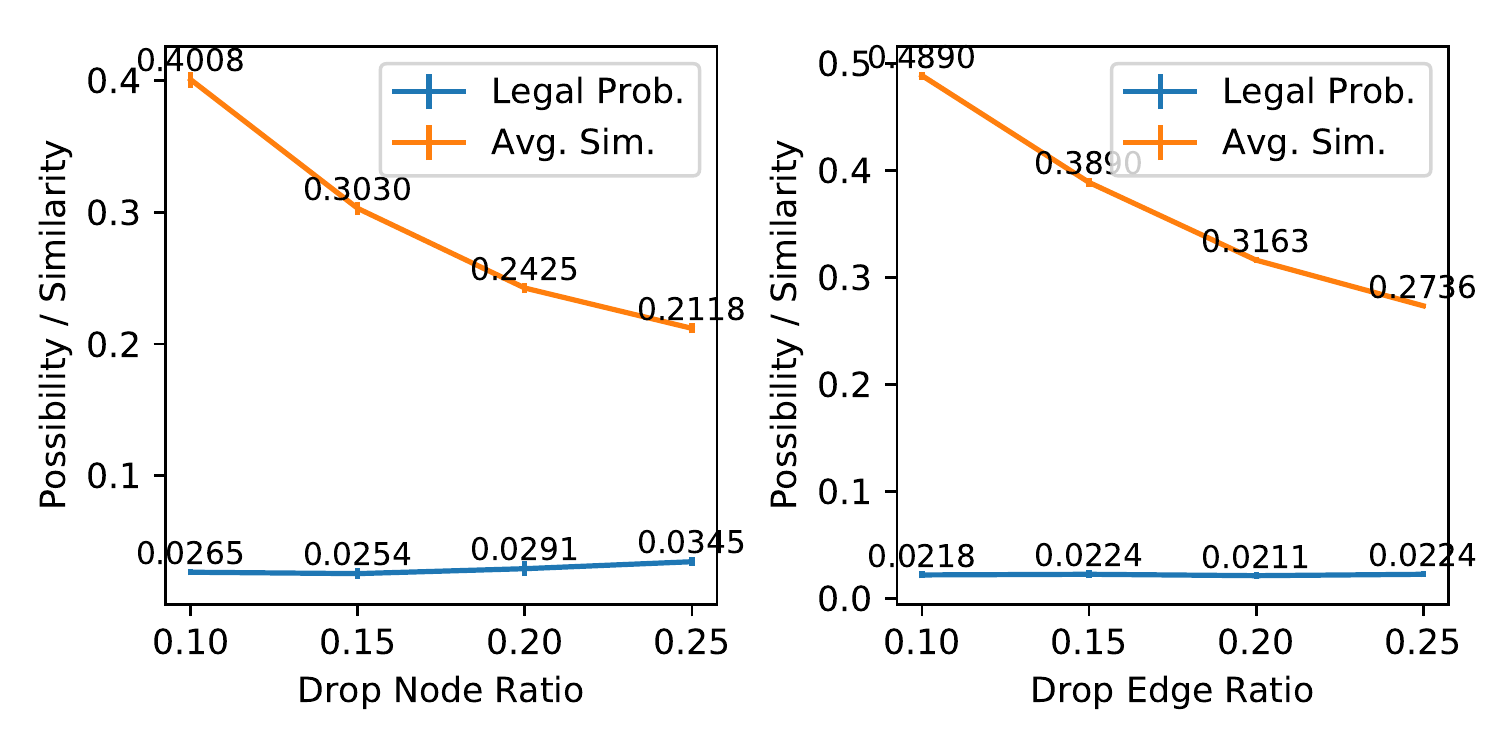}
  \vspace{-4.5ex}
  \caption{\footnotesize The fingerprint similarity scores (in orange) and percent of legal molecular outputs (in blue) generated by graph data augmentation strategies: \emph{dropping nodes} and \emph{dropping edges} from a same molecule w.r.t. the ratio of nodes / edges being dropped on 1000 molecular graphs. The fingerprint similarity and the percent of legal molecular outputs decreased dramatically even for the small proportion of nod/edge dropping. 
  }
  \vspace{-3ex}
  \label{fig_exp_sim_val_drnode_dredge_drprob}
\end{wrapfigure}

We make our assumptions on the \emph{necessary information} that should be preserved in positive instances in the contrastive learning process, which though have not been proved theoretically, are reasonable and are arrived from the re-thinking of the purpose and inherent principle of the contrastive learning and what should positive samples preserve to get an effective method. 
Such unproved but reasonable assumptions for positive instances are as follows: 
\begin{itemize}[leftmargin=.5cm]
    \item Positive instances should be semantically similar with the target instance; 
    \item Positive samples for the same target instance should also be similar with each other;
    \item Positive instances should preserve certain domain information if necessary. 
\end{itemize}
Based on such assumptions, we can see that some widely used graph data augmentation strategies cannot always get positive instances with such properties preserved when being applied on different kinds of graph data. 
As shown in Fig.~\ref{fig_exp_sim_val_drnode_dredge_drprob}, 
simple edge-perturbation or node-dropping for molecular graph contrastive learning strategies can hardly get legal graph instances given the fact that molecules are specifically formulated in accordance to strict chemical constraints which will be easily broken if some edges/nodes, even of a very small number, are perturbed. 
Moreover, subgraph sampling strategy, though effective when applied on graphs without node/edge attributes, may always lead to positive instances that are dissimilar with the target instance when applied on molecular graphs. 
Statistical results for subgraph sampling and another data augmentation strategy suffering from similar problems -- attribute masking, are presented in Appendix~\ref{sec_app_data_augmentation_for_positive_instances}.

Thus, in this paper, we move beyond the widely used graph data augmentation strategies for an effective and more universal method to get positive graph instances for graph instance contrastive learning. 
We propose a simple but effective similarity based positive instances sampling strategy that can be applied on various kinds of graph data. 
Unlike previous methods that construct contrastive pairs by
graph augmentation, our method encodes the pair-wise similarity information, measured by certain domain-specific similarity/proximity, into a hierarchical structure and selects positive graph instances 
from such a structure
which ultimately maintains the legality of the sampled instances and high similarity to the target graphs (see Appendix~\ref{sec_broader_impact} for details). 
Moreover, we also propose an improvement for a widely-used node-level pre-training strategy~\citep{hu2019strategies}, which, together with our similarity aware graph positive sampling strategy, brings us an upper strategy design philosophy. That is, the necessary of introducing prior knowledge or bias in random strategies.

We conduct extensive experiments on three representative kinds of graph data: molecular graphs, social graphs as well as big social and academic graphs where nodes are of interest to demonstrate the effectiveness and superiority of our proposed sampling based strategy over previous graph contrastive learning strategies and also some other strategies not based on contrastive learning for different kinds of graph data. 
Besides, some additional experiments which try to transfer the GNN models pre-trained on molecular graph dataset to downstream social graph classification task let us have a glimpse of the potential possibility of the pre-trained models' ability to capture universal graph structural information underlying different kinds of graph data as well as the possibility to get such a universally transferable pre-trained model. 
Similar things have been explored in other domains such as multi-lingual language models. However, to our best knowledge, we are the first to propose such possibility for pre-trained GNN models, which, though lacks further and thorough exploration in the paper, can probably point out a new possibly meaningful research direction and cast light on successive work.

\section{Related Works} 
\vpara{Graph Representation Learning.}
How to generate expressive representation vectors for nodes or graphs that can capture both node-level information, like node attributes and node proximities~\citep{tang2015line,perozzi2014deepwalk,grover2016node2vec}, as well as graph-level information, like structural proximity between nodes~\citep{ribeiro2017struc2vec} and graph property~\citep{gilmer2017neural}, is a vital question and has aroused great interests from graph learning community. 
Common approaches include unsupervised manners~\citep{perozzi2014deepwalk,grover2016node2vec,tang2015line,sun2019infograph,velickovic2019deep,zhang2019prone,qiu2019netsmf,qiu2018network}, which always adopt a shallow architecture, semi-supervised and supervised approaches~\citep{velivckovic2017graph,kipf2016semi,hamilton2017inductive,gilmer2017neural,xu2018powerful}, which always leverage expressive graph neural networks to capture critical information from both graph structure and node/edge attributes. In this work, we adopt graph neural networks as our graph encoder to generate expressive representations for nodes or graphs. 

\vpara{Contrastive Learning.}
Contrastive learning has proved its efficiency to learn highly expressive representations in Computer Vision domain~\citep{chen2020simple,he2020momentum}. 
Moreover, 
contrastive learning has also been used in graph learning for a long time, like doing contrast between node-node pairs~\citep{perozzi2014deepwalk,grover2016node2vec} to encode various node proximities into node representations. 
Recently, there are also efforts focusing on using contrastive learning on graph instances to learn instance-level representations that can be aware of critical graph structural information~\citep{qiu2020gcc,you2020graph}.  In this work, we also focus on graph instance contrastive learning, but turn to approach this problem in a new manner. 

\vpara{Graph Pre-training.} 
Pre-trained models have proved their highly transferable ability when being applied on downstream datasets in other domains, such as the language models~\citep{devlin2018bert} in NLP domain. 
Famous pre-training strategies for GNNs on graph data largely fall into two genres: node-level and graph-level strategies. 
Node-level strategies aim to design proper tasks that can help GNNs learn node/edge attribute distribution information~\citep{hu2019strategies,rong2020grover}. 
More universally, graph-level strategies try to learn design tasks that can learn structural information for both nodes and graphs~\citep{qiu2020gcc,you2020graph}. In this work, we aim to design more powerful pre-training strategies for graph data from both graph-level and node-level.

\section{Preliminary}
We denote an attributed graph as $G(\mathcal{V}, \mathcal{E}, \mathcal{X})$, where $\left\vert \mathcal{V} \right\vert = n$ refers to a set of $n$ nodes and $\left\vert \mathcal{E} \right\vert = m$ refers to a set of $m$ edges. We denote $\bm{x}_v \in \mathbb{R}^{d}$ as the initial feature of node $v$ and $\bm{e}_{uv}$ as the initial feature of edge $(u,v)$. 

Graph Neural Networks (GNNs) can be modeled as the a messaging passing process, which involves neighborhood aggregation among nodes in graph and message updating to the next layer. Namely, the general message passing process is defined as:
\begin{align}
    \notag\bm{m}^{(l+1)}_v &= \text{AGGREGATE}(\{(\bm{h}_v^{(l)}, \bm{h}_{u}^{(l)}, \bm{e}_{uv}) | u \in \mathcal{N}_v\}),\\
    \notag\bm{h}^{l+1}_v &= \sigma( \bm{W}^{(l)} \bm{m}^{(l+1)}_v + \bm{b}^{(l)} ),
    \label{equ.gnn}
\end{align}
where $\bm{h}^{l+1}_v$ refers to the hidden state of $v$ at $(l+1)$-th layer with $\bm{h}^{(0)}_v = \bm{x}_v$ and $\bm{m}^{(l+1)}_v$ refers to the aggregated message of $v$ at $(l+1)$-th layer.  $\mathcal{N}_v$ denotes the neighbor node set of node $v$. $\text{AGGREGATE}(\cdot)$ aggregates the hidden states of $v$'s neighbor nodes and edges, such as mean/max pooling and graph attention\citep{xu2018powerful,velivckovic2017graph}. $\sigma(\cdot)$ is the activation function, such as $\text{ReLU}(\cdot)$. $\bm{W}^{(l)}$ and $\bm{b}^{(l)}$ are the trainable parameters. 
If the model $\mathcal{M}_{L}$ contains $L$ layers, the output of last layer $\{\bm{h}^{(L)}_v\}_{v \in \mathcal{v}}$ usually represents the node-level embeddings of input graph. Moreover, the graph-level embedding $\bm{h}_G$ is derived by simply applying a $\text{READOUT} $ function as
\begin{align}
    \notag\bm{h}_G = \text{READOUT}(\{\bm{h}^{(L)}_v\}_{v \in \mathcal{N}_v}).
\end{align}
Representations generated by GNNs over graphs, including node-level and graph-level representations, are meaningful embeddings to perform various downstream graph learning tasks, like node classification~\citep{zhu2007combining,bhagat2011node}, graph classification~\citep{sun2019infograph,hu2019strategies,rong2020grover}, and so on. 

\begin{figure}[t] 
  \centering
  
  \includegraphics[width=1\textwidth]{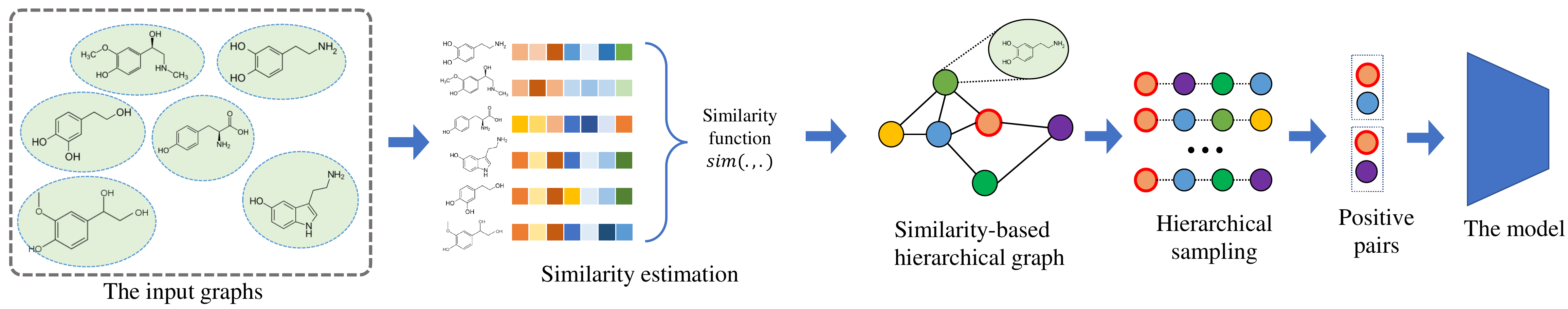}
  \vspace{-2ex}
  \caption{\footnotesize  Illustration of the hierarchical graph instance sampling process for the molecular graphs. 
  }
  \label{fig_model_hier_sampling}
  \vspace{-3ex}
\end{figure}

\section{Similarity-aware Positive Graph Instance Sampling}
In this section, we propose our similarity-aware hierarchical graph positive instance sampling method to sample positive graph instances with three kinds of information mentioned in Sec.~\ref{sec_intro} preserved. 
We first explain our motivation w.r.t. why we turn to other graph instances in the pre-training dataset for positive instances and why it may work for graph data. 
Then we propose our sampling strategies as well as two versions of the sampling process. 
We also give some further discussions for such two sampling strategies. 
Moreover, we also propose an improvement of the widely used node-level pre-training strategy, which is an additional contribution of our work. 

\subsection{Motivation: Sampling or Constructing?} 
As discussed in Sec.~\ref{sec_intro}, it is hard to design a clean and elegant data augmentation strategy universally for various kinds of graph data to get positive instances that are similar enough with the target graph instance and can also preserve necessary domain specific information. 
Since what we care about for positive graph instances are their similarity with the target graph instance, rather than the way to obtaining them, we move beyond popular graph data augmentation skills and propose to sample positive instances from the pre-training dataset for the target graph instance. 
Specifically, we propose to use approximate similarity functions that can reveal the semantic similarity between two graph instances to some extend to estimate the semantic similarity scores between two graph instances. 
The similarity relations between each pair of graphs are then encoded into a similarity hierarchy, which is then used for positive instance sampling. 
We also make some further discussions for the proposed similarity-aware sampling process, which may inspire future design for other sampling strategies.

\subsection{Similarity-aware Positive Graph Instance Sampling 
}
\label{sec_hgc_method_similarity_aware}
Following~\citep{arora2019theoretical}, we assume that each graph instance $G_i\in \mathcal{G}$ has its semantic class $\text{class}(G_i) = c_i$. Thus, the optimal positive sampling strategy should choose graph instances of the same semantic class with the graph instance $G_i$ as its positive instances. Formally, the rate for sampling graph $G_j$ as the positive instance of $G_i$ is: 
\begin{equation}
    P_i^+(G_j) = 
    \begin{cases}
        \frac{1}{\vert \mathcal{G}_i^+\vert } & \text{If class($G_i$) = class($G_j$)}, \\ 
        0 & \text{otherwise}
    \end{cases}
\end{equation}
where $\mathcal{G}_i^+ = \{ G_k | G_k \in \mathcal{G}, \text{class}(G_k) = \text{class}(G_i) \}$ is the set of graph instances of the same class with graph instance $G_i$. 
We can then assume that there exists a ground-truth semantic similarity function $\text{sim}_{\text{gt}}(\cdot, \cdot)$ which reveals whether two graph instances belong to a same semantic class accurately: 
\begin{equation}
    \text{sim}_{\text{gt}}(G_i, G_j) = 
    \begin{cases}
        1 & \text{If class($G_i$) = class($G_j$)}, \\ 
        0 & \text{otherwise}
    \end{cases}
\end{equation}

However, we have no knowledge of the such ground-truth semantic similarity function since our pre-training graph datasets are always unlabeled. 
Thus we propose to use approximate similarity functions that can be obtained from the real-world and applied in practice easily to 
estimate the similarity between two graph instances. 
We can make some assumptions for the chosen approximate similarity functions to ensure their good quality, which are deferred to Appendix~\ref{sec_app_assumption_sim_functions}.

Specifically, we choose a similarity score function $\text{sim}(\cdot, \cdot)$ to estimate the semantic similarity between two graphs. 
To further use the similarity measurement to perform flexible positive sampling, we propose a two-step approach\footnote{Please refer to Appendix~\ref{sec_appen_sel_candi} for details.} to encode pair-wise similarity into a more abstract and structural hierarchy efficiently -- a similarity-based hierarchical graph $\mathcal{H}(\mathcal{G}, \mathcal{E}_H)$, where $\mathcal{G}$ is the set of graphs in our pre-training dataset, $\mathcal{E}_H$ is the edge set. 
Formally, we introduce a similarity threshold $\tau(0 < \tau < 1)$, and based on which the edge set is defined as:  
    $\mathcal{E}_{H} = \{(G_i, G_j) \vert \text{sim}(G_i, G_j) \geq \tau, G_i \in \mathcal{G}, G_j \in \mathcal{G}\}$. 
Many similarity functions are good candidates for $\text{sim}(\cdot, \cdot)$ such as fingerprint similarity~\citep{rogers2010extended} for molecular graphs, Weisfeiler-Lehman Graph Kernel~\citep{shervashidze2011weisfeiler} normalized similarity for graphs without node/edge attributes and node proximity for nodes in a big graph. 

The constructed hierarchical graph, which encodes more information beyond pair-wise similarity\footnote{Such information will be discussed in Sec.~\ref{sec_further_understanding_sim_aware_sampling}.}, can be used to design flexible sampling strategies for positive graph instance selection. 
We propose two sampling strategies: 
\begin{itemize}[noitemsep,topsep=0pt,parsep=0pt,partopsep=0pt,leftmargin=.5cm]
    \item \textbf{First-order neighbourhood sampling.} For each graph $G_i$, sample a one-hop neighbour set of a fixed size 
    as its positive instances. 
    \item \textbf{High-order graph sampling.} We perform $l$-hops random walks starting from graph $G_i$ for $k$ times and choose positive instances according to their appearance frequencies. 
\end{itemize}
An illustration for \hgc\ is presented in Fig.~\ref{fig_model_hier_sampling}.
We will give some further discussions w.r.t. why we use similarity for positive instance sampling and how would high-order sampling potentially benefit the sampling process and the resulting positive instances in the next section. 

\subsection{Further Discussion for Similarity-aware Sampling Strategy} \label{sec_further_understanding_sim_aware_sampling}



In this section, we want to answer two questions: \textbf{Q1:} Why we still sample positive instances based on approximate pair-wise similarity scores, though it may not be an accurate similarity estimation? \textbf{Q2:} How would high-order sampling potentially benefit the sampling process and the resulting positive instances? 
Moreover, we also propose some further discussions for the proposed similarity-aware positive instance sampling strategy. 

To begin with, we propose a property of the contrastive learning that is intuitively correct: 
\begin{property}
    Avoiding false-positives is important in the contrastive learning process.
    \label{claim_false_pos_important}
\end{property}
Here, ``false-positives'' denotes positive instances selected by a non-optimal positive sampling strategy whose semantic classes are not same with the target graph instance. 
We explain why such a property holds in Appendix~\ref{sec_avoid_false_positives} in detail, though it should be correct intuitively. 

Then, \textbf{Q1} can be answered by proposing the following property of the positive instances sampled according to their similarity scores with the target graph instance: 
\begin{property}
    If the similarity threshold $\tau$ is changing in a proper range, 
    an instance that has a high similarity score with the target instance will also has a high probability to be a ground-truth positive instance. 
\end{property}
We would explain why this property holds in detail in Appendix~\ref{sec_app_high_sim_high_gt_pos}, based on our assumptions on good properties of the approximate similarity function (Def.~\ref{def_approximation_similarity_function}).
Thus, the answer for \textbf{Q1} could be: \emph{sampling positive instances according to their similarity scores with the target graph instance may help avoid sampling false-positives.}

To answer \textbf{Q2}, we first propose one limitation of the first-order similarity sampling strategy by pointing out a crucial property of the ground-truth similarity function that the approximate similarity functions always fail to preserve -- \emph{the transitivity of the ground-truth similarity function}: 
\begin{property}[Transitivity of the ground-truth similarity function]
    Ground-truth similarity function is transitive: if $\text{sim}_{\text{gt}}(G_i, G_j) = 1$ and $\text{sim}_{\text{gt}}(G_i, G_k) = 1$, then $\text{sim}_{\text{gt}}(G_j, G_k) = 1$.
\end{property}
Such transitivity of the ground-truth similarity function ensures the transitivity of the relations between nodes in the hierarchical graph constructed based on the ground-truth similarity measurement. 
However, it is obvious that relations between nodes in our constructed similarity-based hierarchical graph -- represented by edges, are not fully-transitive. 
It is because that the approximate similarity function we use in practice is not an optimal one. 

We introduce the definition of connectivity and connectivity order between nodes in the graph in Appendix~\ref{sec_preference_of_ho_sampling}. 
The transitivity of the ground-truth similarity function ensures that $G_i$'s positive instances sampled by first-order neighbourhood sampling strategy can have connectivity orders with $G_i$ ranging from $1$ to $\vert \mathcal{G}_i^+ \vert - 1$. 
It is hard for first-order sampling strategy applied on the hierarchical graph constructed in practice to get positive instances that also have high-order connectivity (e.g., second-order connectivity) with the target graph instance. 
The reason is that first-order information cannot reveal high-order information (e.g., high-order connectivity with the target graph instance) in the constructed hierarchical graph, while it can fully reveal higher-order connectivity in the constructed hierarchical graph based on ground-truth similarity function 
(i.e., if a graph instance $G_j$ is 1-connected to $G_i$, then it is $2,3,...,\vert \mathcal{G}_i^+ \vert-1$ connected to $G_i$ as well). 

We can prove that first-order neighbouring positive instances sampled by second-order sampling process are more likely to be connected with each other (see Appendix~\ref{sec_preference_of_ho_sampling} for details). 
This can remedy the limitation of the first-order sampling strategy, which cannot guarantee the similarity between positive instances. 
Moreover, it can be empirically verified that positive instances that are both first-order and second-order connected to the target instance are also more similar with the target instance. More details are deferred to Appendix~\ref{sec_preference_of_ho_sampling}. 
We also expect that higher-order sampling process can bring more benefit to the resulting positive instances and worth trying in practice. 


Additionally, we propose further discussions w.r.t. how would the changing similarity threshold $\tau$ influence the balance between the increasing sampling rate estimation accuracy for ground-truth positive instances and the risk of sampling more false-positive instances. 
Detailed discussions are deferred to Appendix~\ref{sec_balance_pos_neg}.

\subsection{Adaptive Masking for Node-level Pre-training} 
\label{sec_mtd_adam}
In this section, we propose our improvement of the widely used \emph{attribute masking} node-level pre-training strategy: Adaptive Masking, which is designed for attributed graphs only.  As introduced in~\citep{hu2019strategies}, \emph{attribute masking task}, which is inspired from ``masked language model'' (MLM) in NLP, helps the model learn node/edge attribute distribution across the graph. Formally, attribute masking task is defined as:

\begin{definition}{(Attribute masking task):}\label{def_attr_masking}
Given an attributed graph $G(\mathcal{V}, \mathcal{E}, \mathcal{X})$, a target node $v \in \mathcal{V}$ and its corresponding feature vector $\bm{x}_v$, attribute masking task is first to mask a subset of the features $\bm{x}_{\text{sub}} \subseteq \bm{x}_v$ in feature vector $\bm{x}_v$ 
and produce a new feature vector $\bm{x}_v'$ for node $v$. Then  
let a model $\mathcal{M}$ to make the prediction of the masked feature set $\bm{x}_{\text{sub}}$ given the new feature vector $\bm{x}_v'$ as input. 
\end{definition}

Hu et al.~\citep{hu2019strategies} follows the same protocol in MLM by uniformly selecting the nodes set from graphs to construct the attribute mask task. 
But, we argue that the uniform selection may break structural relations among nodes in graphs so that the model may miss critical information for node attribute distribution from such relations. 
We introduce a toy example in the Appendix~\ref{sec_app_details_adaptive_masking}. 
Inspired by 
Kmean++\citep{arthur2006k}, which aims to obtain the good initial centroids with widely separated in space, we also adopt the adaptive masking (\amn) to generate the mask node set within less correlations.  
In particular, we divide the masking process into $T$ steps. At the first step, we uniformly sample a small mask set. Secondly, the masking weight of each candidate node is adaptive by function $\text{PScore}$.  
The detail of $\text{PScore}$ is demonstrate in Algorithm~\ref{alg:pretubs} (see Appendix~\ref{sec_app_details_adaptive_masking}). 
In $\text{PScore}$, for the candidate node $v$, we calculate the similarity of model output between before and after masking. High similarity indicates that node $v$ is not influenced by the mask operation at the current step, resulting in the low correlation between node $v$ and current mask set $S_{\text{cur}}$. Finally, we randomly sample a node set $\mathcal{K}$  with the probability constructed by masking weight. The algorithmic details are provided in the supplementary material.

According to the adaptive masking operation, we can dynamically adjust the importance of nodes during training and obtain a more representative mask node set for the attribute masking task. 
Such intuition is further discussed in Appendix~\ref{sec_app_effectiveness_adam}.

\section{Experiments}

\subsection{Experimental Configuration} \label{sec_exp_config}


\vpara{Pretraining Data Collection.} 
We conduct the pretraining on four datasets from various domains: 
1). \emph{academic and purchasing graphs}: we collect four data sources from Deep Graph Library \citep{wang2019dgl} and merge them into one pretraining dataset dubbed \text{AP\_NF}. 2). \emph{social graphs}: we construct two pretraining datasets termed \BSGD\ and \SGCD. \BSGD\ contains five data sources, while \SGCD\ contains 13 data sources collected from TUDataset~\citep{Morris+2020}. 3). \emph{molecular graphs}: we use the same  pretraining dataset with 2 million molecules in~\citep{hu2019strategies} and denote it as \text{MolD}. The suffix \text{NF} indicates ``no feature''. Since the data sources have different features, we remove all feature and only pretrain these datasets with \text{HGC}. 
The details are presented in Appendix~\ref{sec_appen_datasets_pret_down_split}.

\vpara{Downstream Tasks.}
We mainly evaluate the peformance on two tasks, node classfication and graph classification. 
For the node classification, we conduct the experiments on two datasets, US-Airport~\citep{ribeiro2017struc2vec} and H-index~\citep{zhang2019oag} following the same splitting protocol in~\citep{qiu2020gcc}. For the graph classification, we conduct the experiments on 11 datasets from molecular graph (7 datasets from \citep{wu2018moleculenet}) and social graphs (4 datasets from \citep{yanardag2015deep}). Details of those datasets are deferred to Appendix~\ref{sec_appen_datasets_pret_down_split}.

\vpara{Baselines. }
For molecular graph classification, we comprehensively compare our pre-training strategies with recent $6$ self-supervised learning strategies for graphs. Among them, Edge\_Pred, Infomax, Attr\_Mask, Context\_Pred, are proposed in~\citep{hu2019strategies}, all of which are node-level pre-training strategies. 
GraphCL~\citep{you2020graph} and C\_Subgraph~\citep{qiu2020gcc} are graph level contrastive pre-training strategies. For node classification and social network graph classification, we compare our model with the best result of GCC~\citep{qiu2020gcc} and several other models (i.e., ProNE~\citep{zhang2019prone}, GraphWave~\citep{donnat2018learning}, DGK~\citep{yanardag2015deep}, graph2vec~\citep{narayanan2017graph2vec}, InfoGraph~\citep{sun2019infograph}, DGCNN~\citep{zhang2018end} and GIN~\citep{xu2018powerful}). 
Details for the implementation, pre-training and fine-tuning settings of baseline models will be discussed in the  Appendix~\ref{sec_appen_imp} and~\ref{sec_app_baselines}.

\vpara{Pre-training Settings.}
We use Adam~\citep{kingma2014adam} for optimization with the learning rate of $0.001$, $\beta_1 = 0.9, \beta_2 = 0.999$ and weight decay of $0$, learning rate warms up over the first $10\%$ steps and then decays linearly. Gradient norm clipping is applied with range $[-1, 1]$. The temperature $\tau$ is set to $0.07$ in \hgc\ pre-training stage. The batch size of \MolD\ pre-training is $256$. For \SGCD\ and \BSGD\ pre-training, the batch size is $32$. For the graph classification task, we use {mean-pooling} to get graph-level representations following~\citep{hu2019strategies}. More pre-training details, including backbones, hyper-parameters and training steps are deferred to Appendix~\ref{sec_appen_fint_config}. 

\vpara{Fine-tuning Settings.}
For each fine-tuning task, we train models for $100$ epochs. For graph classification tasks (whether social graphs or molecular graphs), we select the best model by their corresponding validation metrics, while the last model after $100$ epochs training on downstream training sets are used for further evaluation on downstream evaluation sets, the same with~\citep{qiu2020gcc}. 
We adopt micro F1-score and ROC-AUC as the evaluation measures for different tasks. For molecular dataset, as suggested by~\cite{wu2018moleculenet}, 
we apply three independent randomly initialized runs on each dataset and report the mean and standard deviation.  More details are are deferred to Appendix~\ref{sec_appen_fint_config}.  

\subsection{Results of Downstream Tasks} \label{sec_exp_res}

\begin{table*}[t]
    \centering
    \caption{\footnotesize Experimental results (ROC-AUC) on molecular datasets. The numbers in brackets are standard deviations. 
    Numbers in gray are the best results achieved by backbone models. 
    Bold numbers represent the best results by different backbones. 
    Bold numbers in green represent the best results over all backbones. 
    } 
    \resizebox{0.99\textwidth}{!}{%
    \begin{tabular}{@{\;}c@{\;}|c|c|c|c|c|c|c|c@{\;}}
    \midrule
        \hline
        
        Backbone & Strategy & \textbf{SIDER} & \textbf{ClinTox} & \textbf{BACE} & \textbf{HIV} & \textbf{BBBP} & \textbf{Tox21} & \textbf{ToxCast} \\ \cline{1-9} 
        \specialrule{0em}{1pt}{0pt}
        \multicolumn{2}{c|}{{\#Molecules}} & 1427 & 1478 & 1513 & 41127 & 2039 & 7831 & 8575 \\ 
        \cline{1-9} 
        \specialrule{0em}{1pt}{0pt} 
        \multicolumn{2}{c|}{{\#Prediction tasks}} & 27 & 2 & 1 & 1 & 1 & 12 & 617 \\ 
        \cline{1-9} 
        \specialrule{0em}{1pt}{0pt}
        
        
        
        
        \multirow{9}*{GIN} & GraphCL & $0.5946_{(0.0055)}$ & ${0.6592}_{(0.0074)}$ & $0.7713_{(0.0057)}$ & ${0.7754}_{(0.0093)}$ & $0.7050_{(0.0012)}$ & $0.7562_{(0.0024)}$ & $0.6289_{(0.0023)}$ \\ \cline{2-9} 
        \specialrule{0em}{1pt}{0pt}
        
        ~ & C\_Subgraph & $0.5838_{(0.0022)}$ & $0.6390_{(0.0071)}$ & $0.7736_{(0.0140)}$ & $0.7341_{(0.0079)}$ & $0.6901_{(0.0026)}$ & $0.7521_{(0.0044)}$ & $0.6263_{(0.0061)}$ \\ \cline{2-9} 
        
        ~ & Edge\_Pred & $0.5949_{(0.0032)}$ & $0.6335_{(0.0168)}$ & $0.7939_{(0.0064)}$ & $0.7757_{(0.0096)}$ & $0.6623_{(0.0229)}$ & $0.7589_{(0.0033)}$ & \cellcolor{mygray}${0.6456}_{(0.0023)}$ \\ \cline{2-9}
        \specialrule{0em}{1pt}{0pt}
        
        ~ & Infomax & $0.5755_{(0.0024)}$ & \cellcolor{mygray}${0.6944}_{(0.0187)}$ & $0.7571_{(0.0094)}$ & $0.7653_{(0.0040)}$ & $0.6929_{(0.0054)}$ & $0.7674_{(0.0020)}$ & $0.6302_{(0.0007)}$ \\ \cline{2-9} 
        \specialrule{0em}{1pt}{0pt}
        
        ~ & Attr\_Mask & $0.5947_{(0.0083)}$ & $0.6685_{(0.0093)}$ &  \cellcolor{mygray}${0.8064}_{(0.0042)}$ & $0.7668_{(0.0106)}$ & $0.6316_{(0.0007)}$ & $0.7657_{(0.0054)}$ & $0.6463_{(0.0029)}$  \\ \cline{2-9} 
        \specialrule{0em}{1pt}{0pt}
        
        ~ & Context\_Pred & \cellcolor{mygray}${0.6132}_{(0.0050)}$ & $0.6476_{(0.0168)}$ &  $0.8055_{(0.0115)}$ &  \cellcolor{mygray}${0.7807}_{(0.0054)}$ &  \cellcolor{mygray}${0.7026}_{(0.0097)}$ & \cellcolor{mygray}${0.7715}_{(0.0022)}$ & $0.6427_{(0.0024)}$ \\ \cline{2-9} 
        \specialrule{0em}{1pt}{0pt}
        
        \cline{2-9} 
        
        ~ & \hgc & \cellcolor{green!20}$\mathbf{0.6333}_{(0.0121)}$ & $\mathbf{0.8134}_{(0.0115)}$ & \cellcolor{green!20}$\mathbf{0.8442}_{(0.0138)}$ & $\mathbf{0.7853}_{(0.0072)}$ & ${0.7217}_{(0.0042)}$ & \cellcolor{green!20}$\mathbf{0.7770}_{(0.0022)}$ & $0.6520_{(0.0052)}$ \\ \cline{2-9} 
        \specialrule{0em}{1pt}{0pt}
        
        ~ & \am & $0.6164_{(0.0051)}$ & $0.7797_{(0.0040)}$ & $0.8224_{(0.0041)}$ & $0.7704_{(0.0073)}$ & \cellcolor{green!20}$\mathbf{0.7273}_{(0.0146)}$ & $0.7696_{(0.0014)}$ & \cellcolor{green!20}$\mathbf{0.6603}_{(0.0004)}$  \\ \cline{2-9} 
        \specialrule{0em}{1pt}{0pt} 
        
        ~ & \HGCPAM & $0.6183_{(0.0063)}$ & ${0.7845}_{(0.0499)}$ & $0.8428_{(0.0064)}$ & $0.7839_{(0.0073)}$ & ${0.7172}_{(0.0052)}$  & ${0.7692}_{(0.0030)}$ & $0.6537_{(0.0030)}$ \\ \cline{1-9} 
        \specialrule{0em}{1pt}{0pt}
        
        \multirow{3}*{GCN} & \hgc & $\mathbf{0.6243}_{(0.0044)}$ &  \cellcolor{green!20}$\mathbf{0.8638}_{(0.0051)}$ & $\mathbf{0.8405}_{(0.0006)}$ & ${0.7724}_{(0.0206)}$ & $0.7168_{(0.0014)}$ & ${0.7581}_{(0.0026)}$ & $0.6490_{(0.0024)}$ \\ \cline{2-9} 
        \specialrule{0em}{1pt}{0pt} 
        
        ~ & \am & $0.6209_{(0.0028)}$ & $0.8553_{(0.0044)}$ & $0.8205_{(0.0120)}$ & $0.7693_{(0.0032)}$ & ${0.7018}_{(0.0074)}$ & $0.7533_{(0.0059)}$ & ${0.6449}_{(0.0035)}$ \\ \cline{2-9} 
        \specialrule{0em}{1pt}{0pt}
        
        ~ & \HGCPAM & $0.6164_{(0.0103)}$ & ${0.8231}_{(0.0325)}$ & $0.8249_{(0.0059)}$ & \cellcolor{green!20}$\mathbf{0.7946}_{(0.0102)}$ & $\mathbf{0.7189}_{(0.0103)}$ & $\mathbf{0.7636}_{(0.0070)}$ &  $\mathbf{0.6525}_{(0.0025)}$ \\ \cline{1-9} 
        \specialrule{0em}{1pt}{0pt}
        
        \multirow{3}*{GraphSAGE} & \hgc & $\mathbf{0.6286}_{(0.0016)}$ & ${0.7395}_{(0.0284)}$ & $\mathbf{0.8368}_{(0.0008)}$ & $0.7722_{(0.0149)}$ & $0.7129_{(0.0153)}$ & $0.7583_{(0.0012)}$ & $\mathbf{0.6505}_{(0.0004)}$ \\ \cline{2-9} 
        \specialrule{0em}{1pt}{0pt} 
        
        ~ & \am & $0.6148_{(0.0100)}$ & $0.7098_{(0.0244)}$ & $0.8212_{(0.0019)}$ & $\mathbf{0.7730}_{(0.0057)}$ & ${0.6982}_{(0.0088)}$ & $\mathbf{0.7643}_{(0.0011)}$ & ${0.6492}_{(0.0004)}$ \\ \cline{2-9} 
        \specialrule{0em}{1pt}{0pt}
        
        ~ & \HGCPAM & $0.6250_{(0.0029)}$ & $\mathbf{0.8127}_{(0.0213)}$ & $0.7812_{(0.0038)}$ & $0.7708_{(0.0053)}$ & $\mathbf{0.7187}_{(0.0019)}$ & $0.7610_{(0.0008)}$ & $0.6442_{(0.0018)}$ \\ \cline{1-9} 
        \specialrule{0em}{1pt}{0pt} 

    \end{tabular} 
    }
    \vspace{-4ex}
    \label{tb_mini_prog_modules}
\end{table*}

\subsubsection{Graph Classification}
We evaluate both \hgc\ and \am\ 
on $7$ popular molecular graph classification datasets and \hgc\ 
on $4$ social network graph classification datasets. 

\vpara{The result of molecular graph classification.}
For molecular graph classification datasets, we report our pre-training strategies on different backbones, including GIN~\citep{xu2018powerful}, GCN~\citep{kipf2016semi}, GraphSAGE~\citep{hamilton2017inductive}. Meanwhile, since only \MolD\ contain node features, we apply both \hgc\ and \am\ strategies on the molecular datasets. $\HGCPAM$ indicates the combination of two strategies. 
As shown in Table~\ref{tb_mini_prog_modules}, we have the following observations: \textcolor{blue}{\textbf{(1).}} GIN model pre-trained by our pre-training strategies can consistently outperform those pre-trained by other existing strategies, with large margin on most of them. The overall absolute improvement is 2.98\% in average. 
\textcolor{blue}{\textbf{(2).}} Specially, \hgc\ can consistently outperform those graph-data-augmentation-based contrastive learning strategies (i.e., GraphCL and C\_Subgraph) 
. It verifies our stand point that the graph data augmentation will lose some crucial domain information and compromise the final performance, while \hgc\ dose not lose such information and leads to better performance. 
\begin{wraptable}[9]{r}{0.5\textwidth}
    \vspace{-4ex}
    \caption{\footnotesize  Results on graph classification datasets. The evaluation metric is micro F1-score. 
    }
    \resizebox{0.5\textwidth}{!}{%
\begin{tabular}{l|clcc}
\hline
Strategy                                        & IMDB-B         & IMDB-M         & RDT-B          & RDT-M          \\ \hline
\# graphs                                       & 1000           & 1500           & 2000           & 5000           \\
\# classes                                      & 2              & 3              & 2              & 5              \\ \hline
DGK                                             & 0.670          & 0.446          & 0.780          & 0.413          \\
graph2vec                                       & 0.711          & \textbf{0.504} & 0.758          & 0.479          \\
InfoGraph                                       & 0.730          & 0.497          & 0.825          & 0.535          \\
DGCNN                                           & 0.700          & 0.478          & -              & -              \\
GIN(No-Pret.)                                             & 0.734          & 0.433          & 0.885          & 0.635          \\ \hline
GIN\_GCC (Best)                                      & 0.756          & 0.509          & 0.898      & 0.530          \\
GIN\_\hgc (\BSGD) & \textbf{0.765} & 0.474          & 0.913          & \textbf{0.657} \\
GIN\_\hgc (\SGCD) & 0.756          & 0.490          & \textbf{0.914} & 0.652          \\ \hline
\end{tabular}
    }
    \label{tb_exp_gcc_graph_dataset}
\vspace{-3ex}
\end{wraptable}
\textcolor{blue}{\textbf{(3).}} Even though GCN/GraphSAGE can not surpass the our pre-trained model on GIN pre-trained model, they still outperform the other pretraining strategy, which reaffirms the effectiveness of our pre-training strategies.
\textcolor{blue}{\textbf{(4).}} The combined strategy \HGCPAM\ achieve more benefits on GCN and GraphSAGE than that of GIN. We conjecture that GIN encodes the additional noise which is introduced by this simple combination due to its strong expressive power.

\vpara{The result of social graph classification.}
To check the transferability of \hgc, we conduct the finetune experiments on two  models pretrained by \SGCD\ and \BSGD. \SGCD\ contains the unlabeled data set used in finetune while \BSGD\ does not. Table~\ref{tb_exp_gcc_graph_dataset} documents the performance of GIN model pre-trained by \hgc\ on \SGCD\ and \BSGD\ datasets.  Such results show that GIN model pre-trained by \hgc\ achieves the best performance on three out of four datasets. The comparison between GIN\_\hgc\ and GIN(No-Pret.) also confirms the benefits of \hgc. Another interesting observation is that the pretrain model based on \BSGD\ can obtain the better performance than than \SGCD\ on two out of four datasets. It implies that \hgc\ dose not just memorize the training samples. It can encode the latent structural information from unseen graphs and transfer the knowledge to the downstream tasks.

\subsubsection{Node Classification.}
\begin{wraptable}[8]{r}{0.3\textwidth}
    \vspace{-8ex}
    \caption{\footnotesize Results on node classification datasets. The evaluation metric is  micro F1-score.}
    \resizebox{0.3\textwidth}{!}{%
    \begin{tabular}{@{\;}c@{\;}|c|c@{\;}}
    \midrule
        \hline
        
        Datasets & \textbf{US-Ariport} & \textbf{H-index}  \\ \cline{1-3} 
        \specialrule{0em}{1pt}{0pt}
        
        $|V|$ & 1190 & 5000 \\ 
        
        $|E|$ & 13599 & 44020 \\ \cline{1-3} 
        \specialrule{0em}{1pt}{0pt}
        
        ProNE & 0.623 & 0.691 \\ \cline{1-3} 
        \specialrule{0em}{1pt}{0pt}
        
        GraphWave & 0.602 & 0.703 \\ \cline{1-3} 
        \specialrule{0em}{1pt}{0pt}
        
        Struc2vec & 0.662 & - \\ \cline{1-3} 
        \specialrule{0em}{1pt}{0pt}
        
        GCC (Best) & 0.683 & 0.806 \\ \cline{1-3} 
        \specialrule{0em}{1pt}{0pt}
        
        
        \hgc (\SSAD) & \textbf{0.706} & \textbf{0.824} \\ \cline{1-3} 
        \specialrule{0em}{1pt}{0pt}

    \end{tabular} 
    }
    \label{tb_exp_gcc_node_dataset}
\end{wraptable}
We evaluate our model pre-trained by \hgc\ 
on \SSAD\ 
on two downstream node classification datasets and 
summarize the results 
in Table~\ref{tb_exp_gcc_node_dataset}. 
Among different versions of \GCC, the best ones are presented. 
From Table~\ref{tb_exp_gcc_node_dataset}, the model pre-trained by our \hgc\ strategy can outperform the best \GCC\ model on both datasets. It is worth noting that the pre-training dataset \SSAD\ contains only 70k graphs, which is much smaller than that of \GCC(9M graphs). This verifies the efficiency of \hgc\ in the information extraction.

\subsection{Ablation Study} \label{sec_exp_abl}
\vpara{How useful are the proposed self-supervised tasks?} 

\begin{wraptable}[10]{r}{0.35\textwidth} 
    \vspace{-4ex}
    \caption{\footnotesize Effectiveness of the pre-training on GIN. Bold numbers for absolute improvements larger than $0.05$.}
    \resizebox{0.35\textwidth}{!}{%
    \begin{tabular}{@{\;}c@{\;}|c|c|c@{\;}}
    \midrule
        \hline
        ~ & \textbf{No-Pret.} & \textbf{SS-Pret.} & \textbf{Abs. Imp.} \\ \cline{1-4} 
        \specialrule{0em}{1pt}{0pt}
        
        SIDER & 0.5637 & 0.6333 & \textbf{+0.0696} \\ 

        ClinTox & 0.6480 & 0.8134 & \textbf{+0.1654} \\ 

        BACE & 0.6653 & 0.8442 & \textbf{+0.1789} \\ 

        HIV & 0.7475 & 0.7853 & +0.0378 \\ 

        BBBP & 0.6939 & 0.7273 & {+0.0334} \\ 
        
        Tox21 & 0.7580 & 0.7770 & +0.0190 \\ 

        ToxCast & 0.6370 & 0.6603 & +0.0233 \\ 
        \cline{1-4} 
        \specialrule{0em}{1pt}{0pt}
    \end{tabular} 
    }
    \label{tb_exp_pretrain_no_pretrain_gain}
\end{wraptable}
To evaluate the contribution of our pre-training strategies, we compare the the performance of the pre-trained model by \hgc\ and \am, with the model without any pre-training, each of which shares the same hyper-parameter setting.
Results are summarized in Table~\ref{tb_exp_pretrain_no_pretrain_gain} for backbone GIN. 
It can be seen clearly that all GIN models benefit from self-supervised pre-training tasks on all datasets. 
To be more specific, for GIN, absolute 17.9\% ROC-AUC increase is observed on the dataset BACE, 16.5\% on ClinTox, and 6.96\% on SIDER,  leading to 7.53\% on average. 
Furthermore, pre-trained models gain larger improvement on datasets of relatively small size (e.g., BACE, ClinTox and SIDER), 
which is also observed in~\citep{rong2020grover}. It indicates that self-supervised pre-training helps GNN models learn more inherent graph properties, thus getting better performance in small downstream datasets where labeled graphs are scarce.  
\begin{wraptable}[10]{r}{0.6\textwidth}
    \vspace{-3ex}
    \caption{\footnotesize  Results for pretraining transferability on graph classification datasets.  Numbers in red are the negative transfer cases.
    }
    \resizebox{0.6\textwidth}{!}{%
\begin{tabular}{rr|clcc}
\hline
\multicolumn{1}{l}{\makecell[c]{Pretraining\\Type}} & Strategy                 & \textbf{IMDB-B} & \textbf{IMDB-M} & \textbf{RDT-B}                & \textbf{RDT-M}                         \\ \hline
None                                                & GIN(No-Pret.)            & 0.734           & 0.433           & 0.885                         & 0.635                                  \\ \hline
                                                    & GIN\_GCC (best)          & 0.756           & \textbf{0.509}  & 0.898                         & 0.530                                  \\ \cline{2-6} 
                                                    & \hgc (\BSGD)             & 0.765           & 0.474           & 0.913                         & 0.657                                  \\
\multirow{-3}{*}{Social}                            & \hgc (\SGCD)             & 0.756           & 0.490           & \textbf{0.914}                & 0.652                                  \\ \hline
                                                    & Context\_Pred (\MolD)    & 0.734           & 0.473           & \cellcolor[HTML]{FFCCC9}0.875 & \cellcolor[HTML]{FFCCC9}\textit{0.635} \\
                                                    & S\_Context\_Pred (\MolD) & 0.763           & 0.460           & \cellcolor[HTML]{FFCCC9}0.818 & \cellcolor[HTML]{FFCCC9}0.625          \\
                                                    & \hgc (\MolD)             & \textbf{0.768}  & 0.504           & 0.912                         & 0.656                                  \\
                                                    & \am (\MolD)              & 0.740           & 0.486           & 0.880                         & 0.654                                  \\
\multirow{-5}{*}{Molecular}                         & \HGCPAM (\MolD)          & 0.743           & \textbf{0.509}  & 0.896                         & \textbf{0.665}                         \\ \hline
\end{tabular}
    }
    \label{tb_exp_gcc_graph_dataset_aba_stu}
\end{wraptable} 

\vpara{Can we transfer pre-trained models to downstream datasets that are dramatically different from the pre-training one?  
} \label{sec_exp_abs_transfer} 
It has long been known that the pre-trained model can be generalized to unseen data in pre-training dataset~\citep{qiu2020gcc,hu2019strategies,devlin2018bert,rong2020grover,you2020graph}. 
However, previous literature~\citep{qiu2020gcc,hu2019strategies,you2020graph} largely focuses on transferring the pre-trained model to downstream datasets with similar type of data.  
Here, what we are interested in asking is can we transfer the pre-trained model to the downstream datasets with clearly different type of graphs compared to the ones in the pre-training dataset? 
%
To show this, we demonstrate the case from  \emph{molecular graph}  to \emph{social network graph} classification.
We pre-train GIN in two different ways: one is pretrained by \hgc\ on two social network graph datasets: \BSGD\ and \SGCD, the other is by 
\hgc, \am\ or \HGCPAM\ as well as  Context\_Pred or S\_Context\_Pred~\citep{hu2019strategies} on the molecular dataset \MolD.

The results are summarized in Table~\ref{tb_exp_gcc_graph_dataset_aba_stu},  
which offers the following observations: 
\textcolor{blue}{\textbf{(1).}} Perhaps surprisingly, our methods including \hgc\ and \HGCPAM\ enable the models pre-trained on molecular graphs to even outperform those pre-trained on social graphs. For example, the accuracy of \HGCPAM\ on IMDB-M (0.509) and RDT-M (0.665) is much better than that of \hgc(\BSGD) and \hgc(\SGCD). 
Apart from the universal graph-level properties, the results also inform that larger pre-training datasets can help the model learn such inherent properties better. 
\textcolor{blue}{\textbf{(2).}} Different pre-training strategies could deliver different performance. Models pre-trained by graph-level pre-training strategies or combined strategies (i.e., \hgc(\MolD) and \HGCPAM(\MolD)) can always get better results than those pre-trained by node-level strategies (i.e., \am(\MolD) and Context\_Pred(\MolD)), which indicates that graph-level pre-training strategies can help the model learn global graph-level properties that can be easily transferred to other domains. 
\textcolor{blue}{\textbf{(3).}}  We also observe the \emph{negative transfer} brought by the supervised pretraining in some cases. For instance, S\_Context\_Pred (\MolD)\footnote{Supervised trained by a labeled graph dataset after unsupervised pre-training. See~\citep{hu2019strategies} for details.} get worse performance than its no supervised trained version Context\_Pred (\MolD) on two datasets: RDT-B and RDT-M. It indicates that simple efforts to learn graph-level properties, such as training with labeled graphs, is probable to be limited in the certain domain, thus performing bad in such cross-domain transfer tasks. 
%
Despite this, our \hgc\ and \HGCPAM\ still consistently lead to better performance compared to other pretraining strategies, which, once again, versifies our assumption that our proposed graph contrastive learning strategy can learn more universal, even cross-domain, graph-level patterns. 

\section{Conclusion} 
\label{sec_conclusion}
In this work, we focus on developing an effective, efficient and more universal positive instances sampling method that can be applied on many different kinds of graph data for graph instance contrastive learning. 
We also propose an improvement for a widely used node-level pre-training strategy to adaptively select nodes to mask for an even distribution (\AM). 
Moreover, we also discover the potential cross-domain transferring ability for 
the pre-trained GNN models. 
However, there are still some limitations in our work: 1). Though high-order graph sampling can get positive instances of better quality than those obtained by first-order sampling in our analysis, it cannot always outperform the model pre-trained by first-order sampling process. We guess that it is relevant with the pre-training dataset.  
2). Just combining \hgcn\ and \AM\ in a simple manner leads to no significant improvement. 
Though we make no further investigation into a more effective combination method since it is not the keypoint of the paper, it is a meaningful research direction. 
3). We discover the potential cross-domain transferring ability 
for pre-trained GNN models. It is an interesting point but no further discussion is made in this paper. 
However, further relevant investigation is interesting and meaningful. 

\bibliographystyle{plainnat}
\bibliography{ref}
\clearpage
\section*{Checklist}


\begin{enumerate}

\item For all authors...
\begin{enumerate}
  \item Do the main claims made in the abstract and introduction accurately reflect the paper's contributions and scope?
    \answerYes{The main contribution of this paper is the proposal of an effective and more universal positive instance selection strategy that can be applied on various kinds of graph data in the contrastive learning process. We also propose an improvement of the a widely used node-level pre-training strategy to adaptively choose nodes to make them distributed evenly in the graph. Moreover, we discover the potential possibility of the cross-domain transferable ability of the pre-trained GNN models. }
  \item Did you describe the limitations of your work?
    \answerYes{See Sec.~\ref{sec_conclusion}.}
  \item Did you discuss any potential negative societal impacts of your work?
    \answerYes{See Sec.~\ref{sec_broader_impact}.}
  \item Have you read the ethics review guidelines and ensured that your paper conforms to them?
    \answerYes{}
\end{enumerate}

\item If you are including theoretical results...
\begin{enumerate}
  \item Did you state the full set of assumptions of all theoretical results?
    \answerYes{See Sec.~\ref{sec_app_effectiveness_adam}. We make reasonable assumptions on the possibility density function of the approximate similarity function on the ground-truth positive graph set and negative graph set. Some reasonable approximations are made in the derivation process.}
	\item Did you include complete proofs of all theoretical results?
    \answerYes{See Sec.~\ref{sec_app_effectiveness_adam}.}
\end{enumerate}

\item If you ran experiments...
\begin{enumerate}
  \item Did you include the code, data, and instructions needed to reproduce the main experimental results (either in the supplemental material or as a URL)?
    \answerYes{See Sec.~\ref{sec_appen_datasets_pret_down_split} for descriptions and download links for datasets. Download links for our pre-processed data are shared along with the code. Code is provided with supplemental material. Instructions for reproduction are stated in README.md file in the supplemental material.}
  \item Did you specify all the training details (e.g., data splits, hyperparameters, how they were chosen)?
    \answerYes{See Sec.~\ref{sec_appen_imp} for implementation details, including pre-training and fine-tuning configuration and hyper-parameter selection. See Sec.~\ref{sec_appen_datasets_pret_down_split} for descriptions for datasets and the splitting methods.}
	\item Did you report error bars (e.g., with respect to the random seed after running experiments multiple times)?
    \answerYes{We report mean and std values for 3 independently random initialized run for each evaluation process on molecular graph datasets. See Table~\ref{tb_mini_prog_modules} and Table~\ref{tb_mini_prog_modules_details} for details.}
	\item Did you include the total amount of compute and the type of resources used (e.g., type of GPUs, internal cluster, or cloud provider)?
    \answerYes{See Sec.~\ref{sec_appen_imp} for hardware configurations.}
\end{enumerate}

\item If you are using existing assets (e.g., code, data, models) or curating/releasing new assets...
\begin{enumerate}
  \item If your work uses existing assets, did you citep the creators?
    \answerYes{We provide links for data and code that are from public respiratory we used in our project. We also cite related papers. See Sec.~\ref{sec_appen_datasets_pret_down_split} and Sec.~\ref{sec_app_baselines}.}
  \item Did you mention the license of the assets?
    \answerYes{Datasets obtained from published works are with related papers cited. See Sec.~\ref{sec_appen_datasets_pret_down_split}}
  \item Did you include any new assets either in the supplemental material or as a URL?
    \answerNo{No new datasets are proposed.}
  \item Did you discuss whether and how consent was obtained from people whose data you're using/curating?
    \answerYes{{Download links for public datasets we used are provided. Datasets obtained from published works are with related papers cited. See Sec.~\ref{sec_appen_datasets_pret_down_split}}
  \item Did you discuss whether the data you are using/curating contains personally identifiable information or offensive content?
    \answerYes{Datasets we use are obtained from public datasets, containing no such information. 
    We provide links for them in Sec.~\ref{sec_appen_datasets_pret_down_split}. }
    }. 
\end{enumerate}

\item If you used crowdsourcing or conducted research with human subjects...
\begin{enumerate}
  \item Did you include the full text of instructions given to participants and screenshots, if applicable?
    \answerNA{}
  \item Did you describe any potential participant risks, with links to Institutional Review Board (IRB) approvals, if applicable?
    \answerNA{}
  \item Did you include the estimated hourly wage paid to participants and the total amount spent on participant compensation?
    \answerNA{}
\end{enumerate}

\end{enumerate}


\clearpage
\newpage

\appendix

\section{Experiments}
\subsection{Datasets and downstream tasks}
\label{sec_appen_datasets_pret_down_split}
\vpara{Pre-training datasets.} \SSAD\ is collected from the Deep Graph Library package~\citep{wang2019dgl} with 73832 nodes in total. It is composed of ``computer'', ``photo'' datasets from dgl.data.AmazonCoBuy and ``cs'', ``physics'' datasets from dgl.data.Coauthor. \SGCD\ is collected from TUDataset~\citep{Morris+2020} with 156754 graphs in total, which is composed of REDDIT-MULTI-12K, dblp\_ct1, dblp\_ct2, facebook\_ct1, facebook\_ct2, github\_stargazers, highschool\_ct1, highschool\_ct2, infectious\_ct1, infectious\_ct2, tumblr\_ct1, tumblr\_ct2 and twitch\_egos. \BSGD\ is also collected from TUDataset~\citep{Morris+2020} with 14500 graphs in total, which is composed of IMDB-BINARY, IMDB-MULTI, REDDIT-BINARY, REDDITMULTI-5K and COLLAB\footnote{All datasets from TUDataset can be downloaded from \href{https://chrsmrrs.github.io/datasets/docs/home/}{https://chrsmrrs.github.io/datasets/docs/home/}.}. \MolD\ is composed of 2000000 unlabeled molecular graphs sampled from ZINC15~\citep{sterling2015zinc}, the same pre-training dataset used in~\citep{hu2019strategies}. The pre-training datasets are also summarized in Table~\ref{tb_exp_pretrain_datasets_details}. 

\vpara{Downstream datasets.} 
For downstream evaluation on molecular graphs, we use 7 benchmark datasets from MoleculeNet~\citep{wu2018moleculenet}\footnote{All of those datasets can be downloaded from \href{http://moleculenet.ai/datasets-1}{http://moleculenet.ai/datasets-1}.}. Details of such datasets are presented in Table~\ref{tb_exp_downstream_mol_datasets_details}. For dataset split, we adopt the scaffold splitting~\citep{bemis1996properties} with the ratio for train/validation/test as 8:1:1. It is a more realistic method for molecular property prediction compared with random splitting and is also the one used in~\citep{hu2019strategies,you2020graph}. 
For downstream evaluation on social graph datasets, we use $4$ datasets\footnote{All of them can be downloaded from \href{https://chrsmrrs.github.io/datasets/docs/datasets/}{https://chrsmrrs.github.io/datasets/docs/datasets/}.} from Yanardag and Vishwanathan~\citep{yanardag2015deep}. Details about them are summarized in Table~\ref{tb_exp_downstream_dataset_soc_graphs}. As for dataset splitting method, for each dataset we first split it into train/test sets with the ratio 9:1 and then split the train set into train/validation sets with the ratio 8:1. The validation set is used for model selection. Note that it is different from the splitting method used in \GCC\citep{qiu2020gcc}, where the dataset is randomly split into train/test sets with the ratio 9:1. 
For downstream node classification datasets, we obtain them (i.e., {US-Airport} and {H-index}) from the download link\footnote{\href{https://drive.google.com/open?id=12kmPV3XjVufxbIVNx5BQr-CFM9SmaFvM}{https://drive.google.com/open?id=12kmPV3XjVufxbIVNx5BQr-CFM9SmaFvM}} for the downstream datasets provided by the author of~\citep{qiu2020gcc}. Three different versions of the dataset H-index are provided by the author, among which we use the one named ``rand20intop200\_5000'', which is the same version with the one used in their evaluation process~\citep{qiu2020gcc}.
The way to split those datasets is also kept the same with the one used in~\citep{qiu2020gcc} (i.e., split into train/test with the ratio 9:1 randomly). 

\subsection{Implementation Details} \label{sec_appen_imp}

\begin{table}[h!]
    \centering
    \caption{Detailed hyper-parameter settings in pre-training stage for models with GIN as their backbones on the molecular dataset. For abbreviations used, ``lr'' denotes ``learning rate''; ``PS'' denotes ``Positive samples''; ``NS'' denotes ``negative samples''; 
    \text{HA} is short for \HGCPAM.
    \text{H} is short for \hgcn.}
    \begin{tabular}{@{\;}c@{\;}|c|c|c|c|c@{\;}}
    \midrule
        \hline
        ~ & \am & \text{H} (\FOS) & \text{H} (\HOS) & \text{HA} (\FOS) & \text{HA} (\HOS) \\ \cline{1-6} 
        \specialrule{0em}{1pt}{0pt}
        
        Batch size & 256 & 256 & 256 & 256 & 256 \\
        
        Temperature $\tau$ & - & 0.07 & 0.07 & 0.07 & 0.07 \\
        
        Training steps & 781300 & 156260 & 156260 & 156260 & 156260 \\ 
        
        Warmup steps & 78130 & 15626 & 15626 & 15626 & 15626 \\ 
        
        Initial lr & 0.001 & 0.001 & 0.001 & 0.001 & 0.001 \\ 
        
        \#GNN layers & 5 & 5 & 5 & 5 & 5 \\ 
        
        GNN Hidden size & 300 & 300 & 300 & 300 & 300 \\ 
        
        Weight decay & 0 & 0 & 0 & 0 & 0 \\ 
        
        Adam $\beta_1$ & 0.9 & 0.9 & 0.9 & 0.9 & 0.9 \\ 
        
        Adam $\beta_2$ & 0.999 & 0.999 & 0.999 & 0.999 & 0.999 \\ 
        
        Gradient clipping & 1.0 & 1.0 & 1.0 & 1.0 & 1.0 \\ 
        
        Dropout rate & 0.0 & 0.0 & 0.0 & 0.0 & 0.0 \\ 
        
        Walk length & - & 1 & 2 & 1 & 2 \\ 
        
        \#Walks & - & 1 & 5 & 1 & 5 \\ 
        
        Mask ratio & 0.15 & - & - & 0.15 & 0.15 \\ 
        
        Mask times & 5 & - & - & 3 & 5 \\ 
        
        \#PS & - & 3 & 5 & 3 & 5 \\
        
        \#NS & - & 255 & 255 & 255 & 255 \\
        \cline{1-6} 
        \specialrule{0em}{1pt}{0pt}
    \end{tabular} 
    \label{tb_exp_pretrain_hyper_params_details}
\end{table} 

    \begin{table}[h!]
    \centering
    \caption{Detailed hyper-parameter settings in pre-training stage for models with GCN or GraphSAGE as their backbones on the molecular dataset. Only those which are different from hyper-parameter settings of models using GIN as their backbones are presented.}
    \begin{tabular}{@{\;}c@{\;}|c|c|c|c|c@{\;}}
    \midrule
        \hline
        ~ & \am & \text{H} (\FOS) & \text{H} (\HOS) & \text{HA} (\FOS) & \text{HA} (\HOS) \\ \cline{1-6} 
        \specialrule{0em}{1pt}{0pt}
        Walk length & - & 1 & 4 & 1 & 2 \\ 
        
        \#Walks & - & 1 & 7 & 1 & 5 \\ 
        \cline{1-6} 
        \specialrule{0em}{1pt}{0pt}
    \end{tabular} 
    \label{tb_exp_pretrain_hyper_params_details_other_backbones}
\end{table} 

\begin{table}[h!]
    \centering
    \caption{Hyper-parameter in fine-tuning stage and their search space.}
    \begin{tabular}{@{\;}c@{\;}|c@{\;}}
    \midrule
        \hline
        Hyper-parameter & Range \\ \cline{1-2} 
        \specialrule{0em}{1pt}{0pt}
        
        Learning rate & 0.0001$\sim$0.01 \\
        
        Batch size & 32,64,128,256 \\
        
        Dropout ratio & 0,0.1,0.2,0.3,0.4,0.5,0.6,0.7 \\ 
        
        Learning rate scale & 0.7,0.8,0.9,1.0,1.1,1.2,1.3 \\
        
        Graph pooling method & mean, sum \\ 
        
        Feature combination method & last \\ 
        
        \#Training epochs & 100 \\ 
        
        \cline{1-2} 
        \specialrule{0em}{1pt}{0pt}
    \end{tabular} 
    
    \label{tb_exp_finetuning_hyper_params_details}
\end{table} 

\begin{table}[h!]
    \centering
    \caption{Detailed information of pre-training datasets.}
    \begin{tabular}{@{\;}c@{\;}|c|c@{\;}}
    \midrule
        \hline
        Dataset & \tabincell{c}{\#Graphs\\ /Nodes} & Data Sources \\ \cline{1-3} 
        \specialrule{0em}{1pt}{0pt}
        
        \SSAD & 73832 & \tabincell{c}{(``computers'' and ``photo'') \\ from  dgl.data.AmazonCoBuy,\\ (``cs'' and ``physics'') \\ from  dgl.data.Coauthor} \\
        \cline{1-3} 
        \specialrule{0em}{1pt}{0pt}
        
        \SGCD & 156754 & \tabincell{c}{REDDIT-MULTI-12K, dblp\_ct1\&2, \\ facebook\_ct1\&2, github\_stargazers, \\  highschool\_ct1\&2, infectious\_ct1\&2, \\ tumblr\_ct1\&2, twitch\_egos} \\ 
        \cline{1-3} 
        \specialrule{0em}{1pt}{0pt}
        
        \BSGD & 14500 & \tabincell{c}{IMDB-BINARY, IMDB-MULTI, \\ REDDIT-BINARY, REDDITMULTI-5K, \\ COLLAB} \\ 
        \cline{1-3} 
        \specialrule{0em}{1pt}{0pt}
        
        \MolD & $2000000$ & unlabeled molecules sampled from ZINC15~\citep{sterling2015zinc} \\
        \cline{1-3} 
        \specialrule{0em}{1pt}{0pt}
    \end{tabular} 
    \label{tb_exp_pretrain_datasets_details}
\end{table} 

\begin{table}[h!]
    \centering
    \caption{Detailed information of molecular graph downstream dataset.}
    \begin{tabular}{@{\;}c@{\;}|c|c@{\;}}
    \midrule
        \hline
        Dataset & \#Tasks & \#Compounds \\ \cline{1-3} 
        \specialrule{0em}{1pt}{0pt}
        
        SIDER & 27 & 1427 \\
        \cline{1-3} 
        \specialrule{0em}{1pt}{0pt}
        
        ClinTox & 2 & 1478 \\ 
        \cline{1-3} 
        \specialrule{0em}{1pt}{0pt}
        
        BACE & 1 & 1513 \\ 
        \cline{1-3} 
        \specialrule{0em}{1pt}{0pt}
        
        HIV & 1 & 41127 \\
        \cline{1-3} 
        \specialrule{0em}{1pt}{0pt}
        
        BBBP & 1 & 2039 \\
        \cline{1-3} 
        \specialrule{0em}{1pt}{0pt}
        
        Tox21 & 12 & 7831 \\
        \cline{1-3} 
        \specialrule{0em}{1pt}{0pt}
        
        ToxCast & 617 & 8575 \\
        \cline{1-3} 
        \specialrule{0em}{1pt}{0pt}
    \end{tabular} 
    \label{tb_exp_downstream_mol_datasets_details}
\end{table} 

\begin{table}[h!]
    \centering
    \caption{Detailed information of downstream social graph datasets.}
    \begin{tabular}{@{\;}c@{\;}|c|c@{\;}}
    \midrule
        \hline
        Dataset & \#graphs & \#classes \\ \cline{1-3} 
        \specialrule{0em}{1pt}{0pt}
        
        IMDB-B (IMDB-BINARY) & 1000 & 2 \\
        \cline{1-3} 
        \specialrule{0em}{1pt}{0pt}
        
        IMDB-M (IMDB-MULTI) & 1500 & 3 \\ 
        \cline{1-3} 
        \specialrule{0em}{1pt}{0pt}
        
        RDT-B (REDDIT-BINARY) & 2000 & 2 \\ 
        \cline{1-3} 
        \specialrule{0em}{1pt}{0pt}
        
        RDT-M (REDDIT-MULTI-5K) & 5000 & 5 \\
        \cline{1-3} 
        \specialrule{0em}{1pt}{0pt}
    \end{tabular} 
    \label{tb_exp_downstream_dataset_soc_graphs}
\end{table} 

\begin{table*}[h!]
    \centering
    \caption{Detailed experimental results for different models on molecular datasets. The numbers in brackets are the values of standard deviations. 
    } 
    \resizebox{1\textwidth}{!}{%
    \begin{tabular}{@{\;}c@{\;}|c|c|c|c|c|c|c|c@{\;}}
    \midrule
        \hline
        
        Backbone & Strategy & \textbf{SIDER} & \textbf{ClinTox} & \textbf{BACE} & \textbf{HIV} & \textbf{BBBP} & \textbf{Tox21} & \textbf{ToxCast} \\ \cline{1-9} 
        \specialrule{0em}{1pt}{0pt}
        
        \multirow{4}*{GIN} & \hgc (\FOS) & ${0.6333}_{(0.0121)}$ & $0.7919_{(0.0408)}$ & ${0.8442}_{(0.0138)}$ & ${0.7853}_{(0.0072)}$ & ${0.7217}_{(0.0042)}$ & ${0.7770}_{(0.0022)}$ & $0.6520_{(0.0052)}$ \\ \cline{2-9} 
        \specialrule{0em}{1pt}{0pt}
        
         ~ & \hgc (\HOS) & ${0.6237}_{(0.0077)}$ & ${0.8134}_{(0.0115)}$ & ${0.7982}_{(0.0201)}$ & $0.7687_{(0.0058)}$ & ${0.7200}_{(0.0082)}$ & $0.7622_{(0.0021)}$ & $0.6379_{(0.0066)}$ \\ \cline{2-9} 
        \specialrule{0em}{1pt}{0pt}
        
        ~ & \HGCPAM (\FOS) & $0.6118_{(0.0110)}$ & ${0.7845}_{(0.0499)}$ & $0.8428_{(0.0064)}$ & $0.7839_{(0.0073)}$ & ${0.7118}_{(0.0082)}$ & ${0.7692}_{(0.0030)}$ & $0.6537_{(0.0030)}$ \\ \cline{2-9} 
        \specialrule{0em}{1pt}{0pt}
        
        ~ & \HGCPAM (\HOS) & $0.6183_{(0.0063)}$ & ${0.7281}_{(0.0052)}$ & $0.7927_{(0.0187)}$ & $0.7672_{(0.0113)}$ & ${0.7172}_{(0.0052)}$ & ${0.7635}_{(0.0025)}$ & $0.6459_{(0.0038)}$ \\ \cline{1-9} 
        \specialrule{0em}{1pt}{0pt}

        \multirow{4}*{GCN} & \hgc (\FOS) & ${0.6117}_{(0.0042)}$ & ${0.8638}_{(0.0051)}$ & ${0.8405}_{(0.0006)}$ & ${0.7724}_{(0.0206)}$ & ${0.7047}_{(0.0031)}$ & ${0.7581}_{(0.0026)}$ & $0.6490_{(0.0024)}$ \\ \cline{2-9} 
        \specialrule{0em}{1pt}{0pt}
        
        ~ & \hgc (\HOS) & ${0.6243}_{(0.0044)}$ & $0.8359_{(0.0295)}$ & $0.8000_{(0.0065)}$ & $0.7700_{(0.0029)}$ & $0.7168_{(0.0014)}$ & $0.7552_{(0.0033)}$ & $0.6421_{(0.0022)}$ \\ \cline{2-9} 
        \specialrule{0em}{1pt}{0pt}
        
        ~ & \HGCPAM (\FOS) & $0.6164_{(0.0103)}$ & ${0.8231}_{(0.0325)}$ & $0.8083_{(0.0072)}$ & ${0.7946}_{(0.0102)}$ & ${0.7189}_{(0.0103)}$ & ${0.7636}_{(0.0070)}$ & ${0.6525}_{(0.0025)}$ \\ \cline{2-9} 
        \specialrule{0em}{1pt}{0pt}
        
        ~ & \HGCPAM (\HOS) & $0.6108_{(0.0037)}$ & ${0.7801}_{(0.0313)}$ & $0.8249_{(0.0059)}$ & $0.7701_{(0.0060)}$ & ${0.7006}_{(0.0021)}$ & ${0.7601}_{(0.0017)}$ & $0.6426_{(0.0021)}$ \\ \cline{1-9} 
        \specialrule{0em}{1pt}{0pt}
        
        \multirow{4}*{GraphSAGE} & \hgc (\FOS) & ${0.6286}_{(0.0016)}$ & ${0.7395}_{(0.0284)}$ & ${0.8368}_{(0.0008)}$ & ${0.7583}_{(0.0074)}$ & ${0.7100}_{(0.0016)}$ & ${0.7575}_{(0.0014)}$ & ${0.6505}_{(0.0004)}$ \\ \cline{2-9} 
        \specialrule{0em}{1pt}{0pt}
        
        ~ & \hgc (\HOS) & $0.6130_{(0.0089)}$ & $0.6242_{(0.0466)}$ & $0.7321_{(0.0084)}$ & $0.7722_{(0.0149)}$ & $0.7129_{(0.0153)}$ & $0.7583_{(0.0012)}$ & $0.6379_{(0.0066)}$ \\ \cline{2-9} 
        \specialrule{0em}{1pt}{0pt}
        
        ~ & \HGCPAM (\FOS) & $0.6115_{(0.0040)}$ & $0.7164_{(0.0231)}$ & $0.7741_{(0.0061)}$ & $0.7708_{(0.0053)}$ & ${0.6951}_{(0.0287)}$ & ${0.7379}_{(0.0062)}$ & $0.6423_{(0.0009)}$ \\ \cline{2-9} 
        \specialrule{0em}{1pt}{0pt}
        
        ~ & \HGCPAM (\HOS) & $0.6250_{(0.0029)}$ & ${0.8127}_{(0.0213)}$ & $0.7812_{(0.0038)}$ & ${0.7661}_{(0.0085)}$ & ${0.7187}_{(0.0019)}$ & $0.7610_{(0.0008)}$ & $0.6442_{(0.0018)}$ \\ \cline{1-9} 
        \specialrule{0em}{1pt}{0pt}

    \end{tabular} 
    }
    \label{tb_mini_prog_modules_details}
\end{table*} 

\subsubsection{Pre-training Configuration}  \label{sec_appen_pret_config}
For the fair comparison with other baselines, We use the Graph Isomorphism Network (GIN)~\citep{xu2018powerful} with $5$ layers and $300$ hidden units each layer as our backbones for models pre-trained on all those datasets mentioned above except for \SSAD (whose settings are kept the same with GCC~\citep{qiu2020gcc}), and {mean-pooling} to get graph-level representations following~\citep{hu2019strategies}. 

All the pre-training experiments are conducted on a CentOS server equipped with two Intel(R) Xeon(R) Gold 5120 CPU (2.20GHz) and 504G RAM and 8 NVIDIA 32510MiB GPUs. All models are implemented by PyTorch~\citep{paszke2019pytorch} version 1.4.0, DGL~\citep{wang2019deep} with CUDA version 10.1, PyTorch Geometric~\citep{matthias2019pyg} version 1.4.3, RDKit~\citep{gregxxxrdkit} version 2020.03.2, scikit-learn version 0.22.1 and Python 3.6.10. Information of other packages (like torch\_scatter) is presented with the code provided. 

For \SGCD\ and \BSGD\ pre-training, we train for $171465$ steps with $32$ graphs in each batch. 
For \SSAD\ pre-training, we train for $230800$ steps with $32$ RWR induced subgraphs in each batch and keep other pre-training settings the same with those used in GCC (Moco) pre-training process stated in~\citep{qiu2020gcc}.
For \MolD\ pre-training, \hgc\ and \HGCPAM\ have two versions using those two sampling strategies stated in Section~\ref{sec_hgc_method_similarity_aware} respectively (first-order neighbourhood sampling termed by \FOS\ and high-order graph sampling termed by \HOS). 
Details of the hyper-parameters for different strategies are listed in Table~\ref{tb_exp_pretrain_hyper_params_details} (for models using GIN as their backbones) and Table~\ref{tb_exp_pretrain_hyper_params_details_other_backbones} (for models using GCN or GraphSAGE as their backbones). 
Results for time consumption comparison between \hgcn, \am\ and basic pre-training strategies are presented with the code provided (see ``README.md'' file under the ``supp\_code'' folder). 

\subsubsection{Fine-tuning Configuration} \label{sec_appen_fint_config}
\vpara{Fine-tuning evaluation process details.} 
For node classification tasks using the model pre-trained on \SSAD, we adopt the same settings stated in~\citep{qiu2020gcc} for a fair comparison 
(i.e. Adam~\citep{kingma2014adam} optimizer with learning rate $0.005$, learning rate warms up over the first $3$ epochs, and linearly decays after $3$ epochs). Micro F1-scores on test set after $100$ training epochs are reported. 
For molecular graph classification tasks and social network graph classification tasks, we apply a linear classification layer on top of the pre-trained model, taking pooled graph representations (mean-pooling in our models) as input and output the graph class prediction results. 
we finetune the pre-trained model for $100$ epochs on the train set and report the ROC-AUC (for molecular graph) and  micro F1-score (for social graph) on the test set at the best validation epoch. We apply three independent randomly initialized runs on each dataset and report the mean (and also standard deviation for molecular graph classification). 
For social network classification tasks (whether the model is trained on molecular graph dataset presented in Table~\ref{tb_exp_gcc_graph_dataset_aba_stu} or social network graph dataset presented in both Table~\ref{tb_exp_gcc_graph_dataset_aba_stu} and Table~\ref{tb_exp_gcc_graph_dataset}), we use the Adam optimizer with learning rate $0.001$, weight decay $0$, $\beta_1 = 0.9, \beta_2 = 0.999$, batch size $32$ and learning rate scale for the linear classification layer $1.0$. No hyper-parameter search is performed\footnote{We take the entire graphs as our graph instances in our evaluation process, different from RWR induced subgraphs in the evaluation process of GCC~\citep{qiu2020gcc}.}. We finetune the pre-trained model for $100$ epochs on the train set and report the micro F1-score on the test set at the best validation epoch.
For hyper-parameters in the molecular graph fine-tuning process, we use PBT~\citep{jaderberg2017population} algorithm to search the best combination on the prediction ROC-AUC on the validation set. Details of hyper-parameter search process will be discussed later. 

\vpara{Software and hardware configuration}
Versions of software used in fine-tuning stage are slightly different from those used in the pre-training stage, i.e., Python 3.7.6, Pytorch 1.4.0, DGL 0.5.3, scikit-learn 0.22.1 for the model trained on \SSAD\ fine-tuning and Python 3.6.8, Pytorch 1.5.0, Pytorch Geometric 1.6.1, scikit-learn 0.23.2 for fine-tuning other models.

All the fine-tuning experiments are run on a single P40 GPU. 

\vpara{Hyper-parameter selection.}
For each molecular graph classification task, we use PBT~\citep{jaderberg2017population} algorithm to search for the best hyper-parameter combination on the prediction ROC-AUC on the validation set with the initial trail number set to $10$, parallel trail number $10$, maximum trail number $400$, mix range $0.3$, niche $\sigma$ $0.1$, niche $\alpha$ $1.0$, perturb factor $0.0001$ for continuous parameter (i.e., learning rate). 
Table~\ref{tb_exp_finetuning_hyper_params_details} shows all the hyper-parameters and their search space in the fine-tuning stage.
Search results can be found together with the code provided (see ``README.md'' file under the ``supp\_code'' folder for further instructions). 


\subsection{Baselines} \label{sec_app_baselines}

\subsubsection{Molecular Graph Classification} 

\vpara{Hu. et al.~\citep{hu2019strategies}} Four of the pre-training strategies from Table~\ref{tb_mini_prog_modules} (i.e., Edge\_Pred, Infomax, Attr\_Mask, Context\_Pred) are proposed in~\citep{hu2019strategies}. We download the author's code and pre-trained models that are released officially and apply three independent runs on each downstream fine-tuning dataset using hyper-parameter combinations provided by the authors (i.e., learning rate = $0.001$, dropout rate = $0.5$, batch size = $32$). 

Code: \href{https://github.com/snap-stanford/pretrain-gnns}{https://github.com/snap-stanford/pretrain-gnns}

\vpara{GraphCL~\citep{you2020graph}.} We download the author's code released officially and use the weights of the pre-trained model they provide (i.e., \verb|graphcl_80.pth|). We apply three independent runs on each downstream fine-tuning dataset using the hyper-parameters provided by the authors (i.e., leaning rate = $0.001$, dropout rate = $0.5$, batch size = $32$, number of training epochs = $100$, learning rate scale for the linear classification layer = $1.0$). 

Code: \href{https://github.com/Shen-Lab/GraphCL}{https://github.com/Shen-Lab/GraphCL}

\vpara{C\_Subgraph~\citep{qiu2020gcc}.} We download the code of~\citep{qiu2020gcc} 
released by the authors officially. Since their implementation cannot be used in both our molecular graph pre-training stage as well as the downstream molecular graph classification stage directly, 
we carefully implement their graph sampling strategy, keeping the restart probability and the walk length same with their default settings (i.e., restart probability = $0.8$, walk length is determined by a fixed number and the node's degree, the difference is that we would perform such RWR process for several times to sample enough nodes due to the different version of DGL we use from GCC (see~\ref{sec_appen_pret_config})) 
and use it to pre-train our model using the same pre-training settings with our \hgc\ pre-training stage. The downstream evaluation process is kept the same with those for our model (on molecular graph classification datasets). 

Code: \href{https://github.com/THUDM/GCC}{https://github.com/THUDM/GCC}

\subsubsection{Node Classification \& Social Network Graph Classification}

All the results of the baselines presented in Table~\ref{tb_exp_gcc_node_dataset} and Table~\ref{tb_exp_gcc_graph_dataset} are taken directly from the paper~\citep{qiu2020gcc}, since the test set selection process and the evaluation metric (i.e., micro F1-score) are kept the same with~\citep{qiu2020gcc} 
and that we carefully check the author's description and implementation details of the baselines used in their paper and choose to trust in their implementation and evaluation for those baselines. 

\subsection{Details about Experimental Setup}


\subsubsection{Two-step hierarchical graph construction process}
\label{sec_appen_sel_candi}
We adopt a two-step approaches to construct the hierarchical graph: 
\begin{itemize}[noitemsep,topsep=0pt,parsep=0pt,partopsep=0pt,leftmargin=.5cm]
    \item Candidate selection: We first sort graphs in the dataset by molecular weight (calculated by MolWt($\cdot$) function in the software RDKit~\citep{gregxxxrdkit}) for molecular graphs or number of nodes for graphs without node features. For molecular graphs, we select molecule B as A's candidates if and only if: 1). The molecular weights of A and B differ by no more than 10\% of A; 2). The number of rings of A and B differs by no more than 1; 3). The number of atoms in A and B differs by no more than 7; 4). The number of candidates have been selected is still less than a manually defined value ($70$ in our pre-processing). For graphs without node attributes, we select graph A as graph B's candidate if and only if: 1). The number of nodes of A and B differ by no more than 10\%. 2). The number of edges of A and B differs no more than 10\%. 3). The number of candidates have been selected is still less than a manually defined value ($70$ in our pre-processing).
    \item Edge construction: For each graph A in graph B's candidate set, we calculate their similarity score and build an edge between them in the constructing hierarchical graph if their similarity score is above a pre-defined threshold $\tau$. 
\end{itemize}

\subsubsection{Details of adaptive masking process}
\label{sec_app_details_adaptive_masking}
\vpara{Detailed algorithms.}
The main algorithm of the adaptive masking (\AM) process is summarized in Algorithm~\ref{algo_masking}. 
Details of its sub-algorithm -- $\text{PScore}$ is demonstrated in Algorithm~\ref{alg:pretubs}, where $\text{MASKNODE}(G, \mathcal{S})$ takes a graph $G$ and a node set for masking $\mathcal{S}$ as input and outputs a graph $G'$ with attributes of nodes in $\mathcal{S}$ masked. 

\begin{wrapfigure}[17]{r}{0pt}
\vspace{-2ex}
\includegraphics[width = 0.45\textwidth]{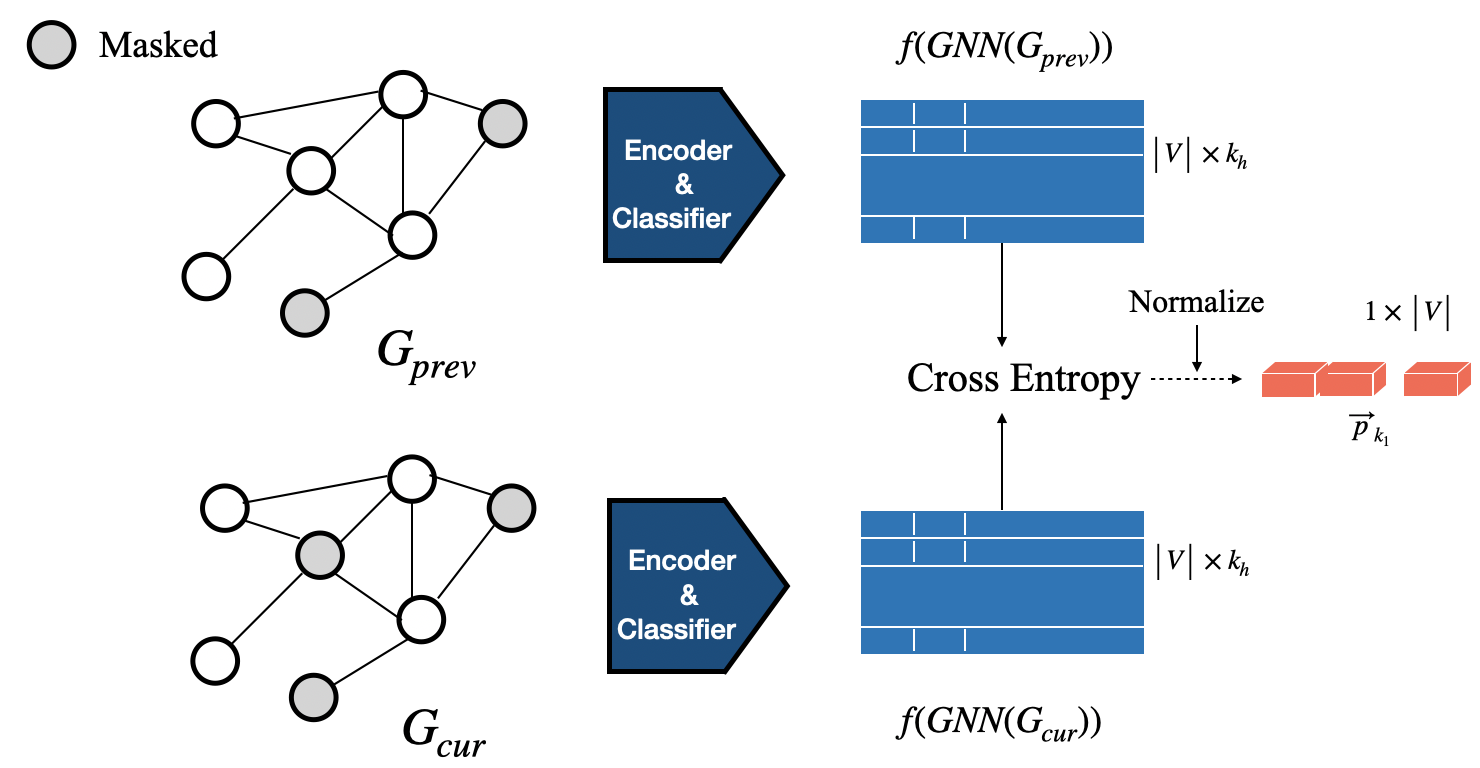}
  \caption{A toy example to illustrate how we determine the masking probability in each masking step (i.e., $k_1$-th), assuming that $k_1 > 2$.}
  \vspace{1ex}
  \label{fig_model_mask_process}
\end{wrapfigure}

\vpara{Examples.}
An example regarding to the adaptive masking process is illustrated in Fig.~\ref{fig_model_mask_process}. 
A toy example mentioned in Sec.~\ref{sec_mtd_adam} claiming that the uniform selection strategy may break structural relations among nodes in graphs is: suppose that node $v$ has only one neighbor: node $u$. In this case, if we mask all features of $u$ and $v$ simultaneously, it is very hard for model $\mathcal{M}$ to make a good prediction of $v$ since $v$ is highly related to $u$. In other words, $\mathcal{M}$ cannot encode the attribute distribution of $v$. Therefore, a good masking set should have less correlations with respective to the output of the model $\mathcal{M}$.

\begin{algorithm}[H]
\small 
\caption{Adaptive Masking. 
}
\label{algo_masking}
\footnotesize
    \begin{algorithmic}[1]
        \Require
            Input graph $G(\mathcal{V}, \mathcal{E}, \mathbf{X})$; The model $\mathcal{M}$;  Masking steps $T$; The number of nodes for masking at each step $\alpha$.
        \Ensure
            Masked node set $\mathcal{S}$; 
        
        \State $\mathcal{S} \leftarrow \emptyset $
        \State $\mathcal{S}_{prev} \leftarrow \emptyset $
        \For{$t = 1$ to $T$}
            \If {t == 1} 
                \State Randomly select a node set $\mathcal{K}$ with $\alpha$ nodes from $\mathcal{V}$ with a uniform distribution. 
            \Else 
                
            \For{$v \in \mathcal{V} \setminus \mathcal{S}$}
            \State $s_v \leftarrow \text{PScore}(v, \mathcal{M}, G, \mathcal{S}, \mathcal{S}_{prev})$
            \EndFor
            \State Randomly select a node set $\mathcal{K}$ with $\alpha$ nodes form $\mathcal{V} \setminus \mathcal{S}$ with probability for each node $v$   : 
            \begin{align}
                \notag p_v = \frac{s_v}{\sum_{u \in \mathcal{V} \setminus \mathcal{S}}s_u}
            \end{align}
            \EndIf 
            \State $\mathcal{S}_{prev} \leftarrow \mathcal{S}$
            \State $\mathcal{S} \leftarrow \mathcal{S} \cup \mathcal{K}$
            
        \EndFor\\
        \Return $\mathcal{S}$
        
    \end{algorithmic}
\end{algorithm}

\begin{algorithm}[H]
\caption{PScore.
}\label{alg:pretubs}
    \begin{algorithmic}[1]
        \Require 
            Target node $v$; 
            The model $\mathcal{M}$;
            Input graph $G(\mathcal{V}, \mathcal{E}, \mathcal{X})$; 
            Current masked node set $\mathcal{S}_{\text{cur}}$;
            Previous masked node set $\mathcal{S}_{\text{prev}}$; 
        \State $G_{\text{prev}} \leftarrow \text{MASKNODE}(G, \mathcal{S}_{\text{prev}})$;
        \State $G_{\text{cur}} \leftarrow \text{MASKNODE}(G, \mathcal{S}_{\text{cur}})$; 
        \State $\bm{y}_{v, \text{prev}} \leftarrow \mathcal{M}(v, G_{\text{prev}})$; 
        \State $\bm{y}_{v, \text{cur}} \leftarrow \mathcal{M}(v, G_{\text{cur}})$;
        \State $s \leftarrow 1 - \text{cross\_entropy}(\bm{y}_{v, \text{prev}}, \bm{y}_{v, \text{cur}})$; \\
        \Return s; 
    \end{algorithmic}
\end{algorithm}

\subsubsection{Time Consumption for Pre-processing}  \label{sec_appen_time_consumption}
No more than $4$ hours for \MolD\ and about $2$ hours for \SGCD, details are presented with code provided (see ``README.md'' file under the ``supp\_code'' folder). 

\subsection{Additional Experimental Results}
\subsubsection{More experimental results for the influence of data augmentations on the quality of positive instances}
\label{sec_app_data_augmentation_for_positive_instances}
We present more experimental results w.r.t. how the quality of resulting positive instances is influenced by data augmentation methods applied. 
This part is to support the claim made in Sec.~\ref{sec_intro} that popular data augmentation strategies cannot get positive graph instances with ideal properties preserved for various kinds of graph data in addition to Fig.~\ref{fig_exp_sim_val_drnode_dredge_drprob}. 

\begin{figure} 
\centering
\includegraphics[width=0.48\textwidth]{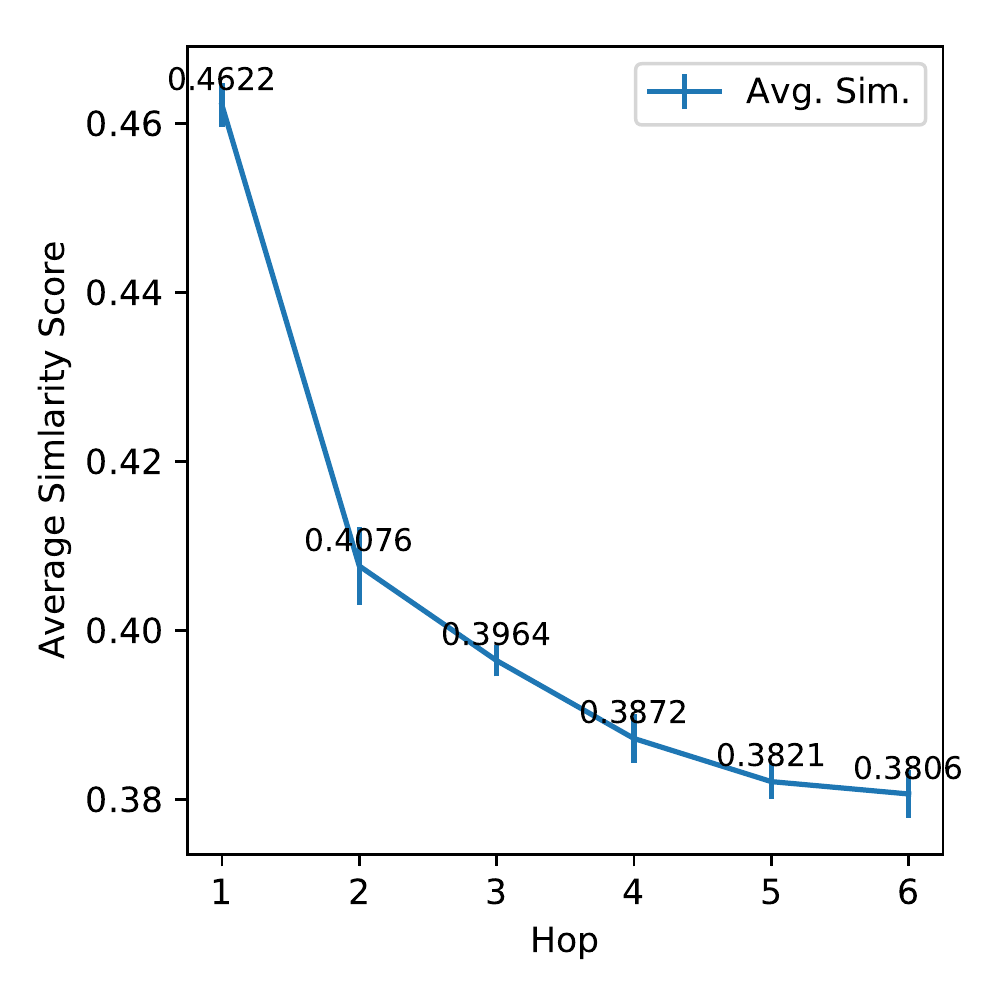}
  \caption{\footnotesize The fingerprint similarity scores between graph instances in the hierarchical graph and the target graph instance w.r.t. the number of hops. Calculated on 1000 randomly selected molecules from \MolD. Three independent runs were given, with average similarity scores and standard deviation reported.
  }
  \vspace{3ex}
  \label{fig_exp_sim_avg_with_hop}
\end{figure}

The claim that the constructed similarity based hierarchical graph encodes the similarity hierarchy is reasonable and can be supported by statistical results for average similarity scores between the target graph instances and their neighbouring graph instances in different hops. 
As shown in Fig.~\ref{fig_exp_sim_avg_with_hop}, the average fingerprint similarity score between molecules in different hops and the target graph instance decreases as the hop increases. 
Moreover, the deceasing speed is relatively low compared with the decreasing speed of the average similarity scores between the positive instances obtained by graph data augmentation strategies and the target graph instance w.r.t. the augmentation ratio (e.g. Fig.~\ref{fig_exp_sim_val_drnode_dredge_drprob}). 

\vpara{Analysis for graph sampling (subgraph) strategy on molecular graphs.}
Graph sampling data augmentation strategy samples subgraphs as positive instances for the target graph instance. 
It is an effective strategy for graphs with no node/edge attributes~\citep{qiu2020gcc}. 
Admittedly, perhaps it can also get subgraphs that are structurally similar enough (measured by Weisfeiler-Lehman Graph Kernel normalized similarity) with each other whether for those graphs with node/edge attributes. It can hardly maintain specific domain information that is important for some kinds of graph data (e.g., molecular graphs) or cannot guarantee enough semantic similarity between obtained positive graph instances and the original graph instance. 
We use the code for subgraph sampling provided by~\citep{you2020graph} to investigate further. We discover that 1). subgraphs cannot preserve enough semantic similarity (measured by fingerprint similarity scores) between obtained positive instances and the target graph instance for molecular graphs; 2). sampled subgraphs may not be proper molecules anymore. 
Statistical results calculated on $1000$ molecules randomly sampled from our molecular pre-training dataset when the subgraph sampling ratio is relatively low ($\le 0.25$) and relatively high (up to $0.8$) are presented in Table~\ref{tb_exp_proper_similarity_ratios_subgraph_sampling} and Table~\ref{tb_exp_proper_similarity_ratios_subgraph_sampling_high} respectively. Perhaps it is surprising that the fingerprint similarity scores between two sampled subgraphs still remain at a low level even when the data augmentation ratio is relatively high (i.e. up to $0.8$). 

\begin{table}
    \centering
    \vspace{-3ex}
    \caption{\footnotesize Statistical results for the ratio of sampled subgraphs to be proper molecules and their similarity scores with the target molecule w.r.t. subgraph sampling ratio. Calculated on $1000$ molecules randomly sampled from \MolD.
    Three independent runs were given, with mean and standard deviation values reported. This table presents results when the subgraph sampling ratio is relatively low.
    Values presented in the table have the format $\text{mean}\pm \text{std}$.}
    \resizebox{1.0\textwidth}{!}{%
    \begin{tabular}{@{\;}c@{\;}|c|c|c|c@{\;}}
    \midrule
        \hline
        ~ & \textbf{0.10} & \textbf{0.15} & \textbf{0.20} & \textbf{0.25} \\ \cline{1-5} 
        \specialrule{0em}{1pt}{0pt}
        Ratio to be proper molecules & $0.511\pm 0.011$ & $0.456\pm 0.005$ & $0.477\pm 0.034$ & $0.522 \pm 0.028$ \\ 
        Similarity scores & $0.00239\pm 0.00008$ & $0.00292\pm 0.00007$ & $0.00296\pm 0.00010$ & $0.00235\pm 0.00003$ \\
        \cline{1-5} 
        \specialrule{0em}{1pt}{0pt}
    \end{tabular} 
    }
    \vspace{-2ex}
    \label{tb_exp_proper_similarity_ratios_subgraph_sampling}
\end{table}

\begin{table}
    \centering
    \vspace{-3ex}
    \caption{\footnotesize Statistical results for the ratio of sampled subgraphs to be proper molecules and their similarity scores with the target molecule w.r.t. subgraph sampling ratio. Calculated on $1000$ molecules randomly sampled from \MolD. Three independent runs were given, with mean and standard deviation values reported. This table presents results when the subgraph sampling ratio is relatively high.
    Values presented in the table have the format $\text{mean}\pm \text{std}$.}
    \resizebox{1.0\textwidth}{!}{%
    \begin{tabular}{@{\;}c@{\;}|c|c|c@{\;}}
    \midrule
        \hline
        ~ & \textbf{0.50} & \textbf{0.70} & \textbf{0.80} \\ \cline{1-4} 
        \specialrule{0em}{1pt}{0pt}
        Ratio to be proper molecules & $0.511\pm 0.012$ & $0.515\pm 0.001$ & $0.506\pm 0.014$ \\ 
        Similarity scores & $0.00211\pm 0.00004$ & $0.00219\pm 0.00008$ & $0.00223\pm 0.00001$ \\
        \cline{1-4} 
        \specialrule{0em}{1pt}{0pt}
    \end{tabular} 
    }
    \vspace{-2ex}
    \label{tb_exp_proper_similarity_ratios_subgraph_sampling_high}
\end{table}

\vpara{Analysis for attribute masking strategy on molecular graphs.}
We also observe similar phenomenon for attribute masking strategy when applied on molecules. 
As shown in Table~\ref{tb_exp_proper_similarity_ratios_attr_masking}, the average similarity score between the resulted positive graph instances and their original graph instance drops as the attribute masking ratio increases, which is not a surprising phenomenon. 
However, unlike the graph sampling strategy, attribute masking strategy always fail to get masked graphs that are legal molecules, perhaps due to the loss of node attributes which are important for molecules. 
\begin{table}
    \vspace{-3ex}
    \caption{\footnotesize Statistical results for the ratio of the resulting masked graphs to be proper molecules and their similarity scores with their respective original molecules w.r.t. attribute masking ratio. Calculated on $1000$ molecules randomly sampled from our pre-training dataset. Three independent runs were given, with mean and standard deviation values reported.}
    \resizebox{1.0\textwidth}{!}{%
    \begin{tabular}{@{\;}c@{\;}|c|c|c|c@{\;}}
    \midrule
        \hline
        ~ & \textbf{0.10} & \textbf{0.15} & \textbf{0.20} & \textbf{0.25} \\ \cline{1-5} 
        \specialrule{0em}{1pt}{0pt}
        Ratio to be proper molecules & $0.014\pm 0.002$ & $0.015\pm 0.003$ & $0.014\pm 0.002$ & $0.014\pm 0.003$\\ 
        Similarity scores & $0.472\pm 0.004$ & $0.414\pm 0.001$ & $0.378\pm 0.003$ & $0.359\pm 0.002$\\
        \cline{1-5} 
        \specialrule{0em}{1pt}{0pt}
    \end{tabular} 
    }
    \label{tb_exp_proper_similarity_ratios_attr_masking}
\end{table}

\begin{figure} 
\centering
\includegraphics[width=0.48\textwidth]{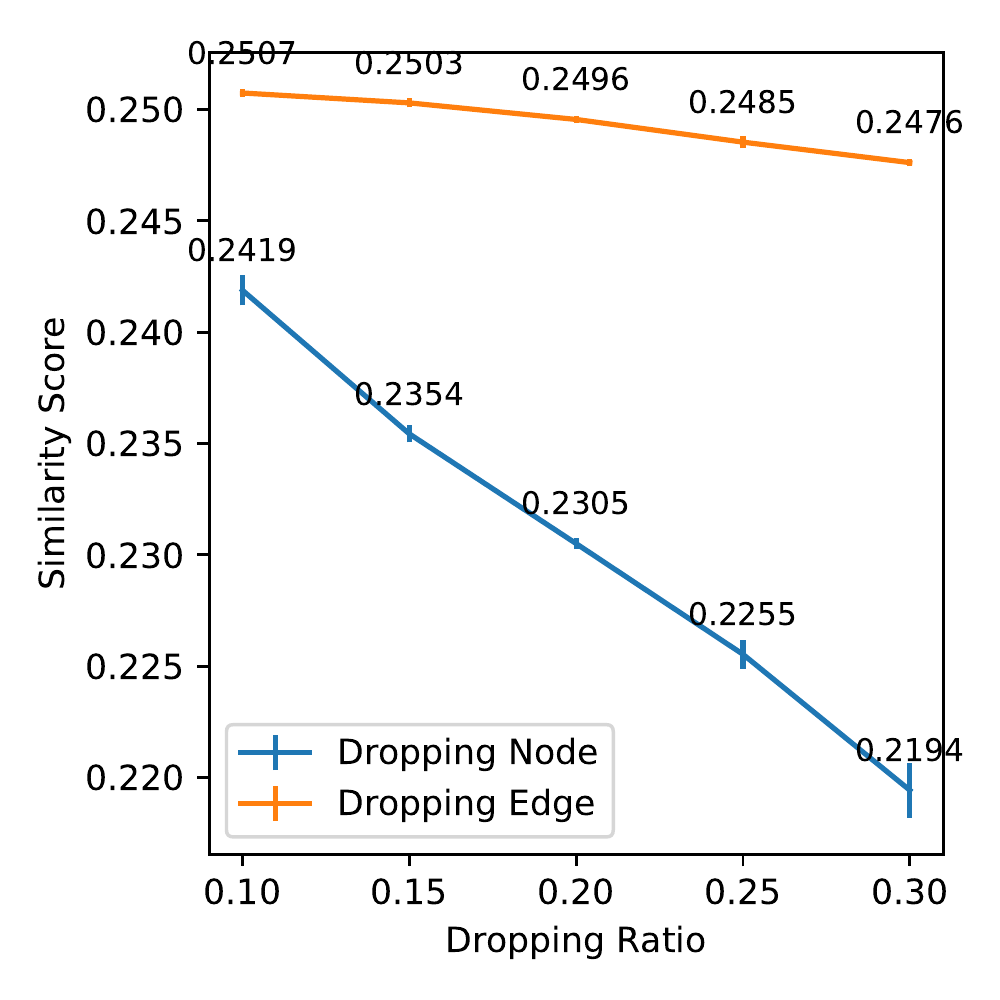}
  \caption{\footnotesize The Weisfeiler-Lehman Graph Kernel normalized similarity scores between positive graph instances, obtained by applying dropping nodes or dropping edges data augmentation strategies, 
  and their respective target graph instances w.r.t. augmentation ratio. Three independent runs were given, with average similarity scores and standard deviation values reported.
  }
  \vspace{-2ex}
  \label{fig_exp_dropping_node_edge_ratio_sim_soc_graphs}
\end{figure}

\begin{figure} 
\centering
\includegraphics[width=0.70\textwidth]{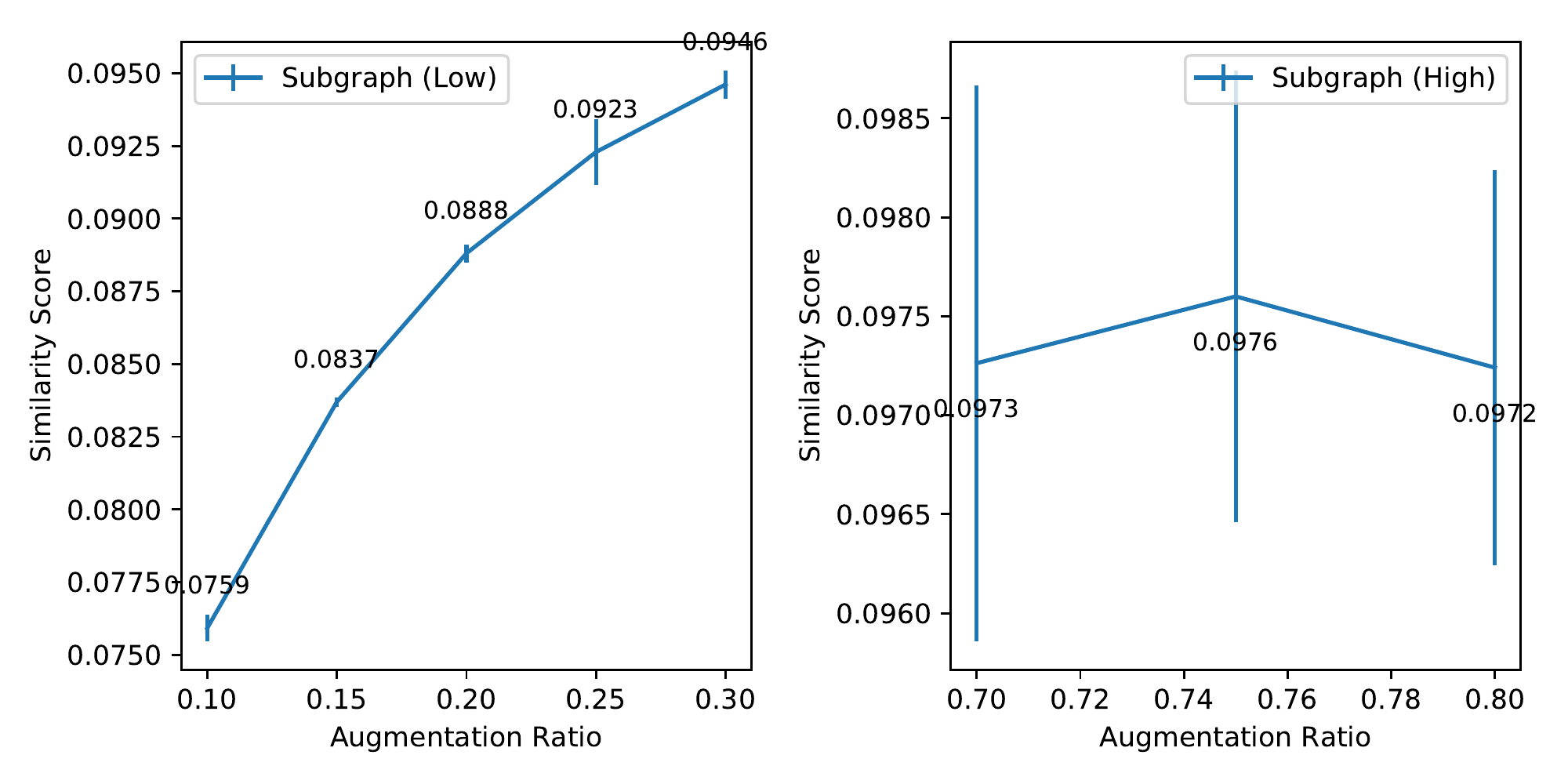}
  \caption{\footnotesize The Weisfeiler-Lehman Graph Kernel normalized similarity scores between positive graph instances, obtained by applying the subgraph data augmentation strategy, and their respective target graph instances 
  w.r.t. subgraph augmentation ratio. Left: changing curve of the average similarity score w.r.t. data augmentation ratio (relatively low).  
  Right: changing curve of the average similarity score w.r.t. data augmentation ratio (relatively high). 
  Three independent runs were given, with average similarity scores and standard deviation values reported.
  }
  \vspace{-2ex}
  \label{fig_exp_subgraph_low_high_sim_soc_graphs}
\end{figure}

\vpara{Analysis for the effectiveness of graph data augmentation strategies on social network graphs.}
We also investigate into how common graph data augmentation strategies will influence the resulting positive instances' similarity scores with the target graph instances when applied on social network graphs.
The results are shown in Fig.~\ref{fig_exp_dropping_node_edge_ratio_sim_soc_graphs} for dropping nodes and dropping edge strategies and Fig.~\ref{fig_exp_subgraph_low_high_sim_soc_graphs} for subgraph augmentation strategy. The experiment is conducted on $1000$ graph instances uniformly randomly chosen from our \SGCD\ dataset. Three independent experiments are performed with the average and standard deviation value of the average similarity score over such $1000$ graph instances in each run are reported. The similarity score measurement is Weisfeiler-Lehman Graph Kernel normalized similarity provided by Python Package GraKel~\citep{Siglidis2020grakel}. Compared with the average similarity between the target graph instance and its first-order graph instances in the constructed hierarchical graph, 
which is $0.3467$, such two data augmentation strategies cannot get positive graph instances that are similar enough with the target instance even when the data augmentation ratio is relatively low (e.g. $0.1$). 
Moreover, the average similarity score drops obviously as the augmentation ratio increases, which is not a surprising phenomenon. 

As for the subgraph augmentation strategy, it can be seen that the average similarity score lies in a low level for both low and high augmentation ratio. 
Such phenomenon is similar with the one observed on molecular graphs. 

\subsection{Additional ablation study of the effectiveness of pre-training strategies on GCN and GraphSAGE}
Similar with the effectiveness of our pre-training strategies (\hgc\ and \am) for GIN models, we can also observe significant contribution of our pre-training strategies on other backbones (GCN and GraphSAGE) as shown in Table~\ref{tb_exp_pretrain_no_pretrain_gain_gcn_sage}. 
We can arrive at a conclusion, which is similar with the one made in~\citep{hu2019strategies}, that more powerful backbone can benefit more from pre-training by comparing Table~\ref{tb_exp_pretrain_no_pretrain_gain_gcn_sage} with Table~\ref{tb_exp_pretrain_no_pretrain_gain}. 
Besides, pre-trained GCN models can get larger benefit on the dataset BACE (27.72\% absolute improvement) compared with the improvement made by pre-trained GIN (16.54\%) and GraphSAGE (10.83\%). 

\begin{table} 
    \centering
    \vspace{-3ex}
    \caption{\footnotesize Effectiveness of the pre-training on GCN and GraphSAGE (denoted as ``SAGE'' in brackets). Bold numbers for absolute improvements larger than $0.05$. For abbreviations used, ``No-Pret.'' denotes trained-from-scratch models, ``Pret.'' denotes pre-trained models (presented are the best values achieved by our different pre-training strategies for each downstream evaluation dataset), ``Abs. Imp.'' denotes the absolute ROC-AUC improvement. }
    \resizebox{1.0\textwidth}{!}{%
    \begin{tabular}{@{\;}c@{\;}|c|c|c|c|c|c@{\;}}
    \midrule
        \hline
        ~ & \textbf{No-Pret. (GCN)} & \textbf{Pret. (GCN)} & \textbf{Abs. Imp. (GCN)} & \textbf{No-Pret. (SAGE)} & \textbf{Pret. (SAGE)} & \textbf{Abs. Imp. (SAGE)} \\ \cline{1-7} 
        \specialrule{0em}{1pt}{0pt}
        
        SIDER & 0.6072 & 0.6243 & +0.0171 & 0.6173 & 0.6286 & +0.0113 \\ 

        ClinTox & 0.5866 & 0.8638 & \textbf{+0.2772} & 0.6936 & 0.8127 & \textbf{+0.1191} \\ 

        BACE & 0.7652 & 0.8405 & \textbf{+0.0753} & 0.7285 & 0.8368 & \textbf{+0.1083} \\ 

        HIV & 0.7547 & 0.7946 & +0.0399 & 0.7477 & 0.7730 & +0.0253 \\ 

        BBBP & 0.6758 & 0.7189 & {+0.0431} & 0.6826 & 0.7187 & +0.0361 \\ 
        
        Tox21 & 0.7370 & 0.7636 & +0.0266 & 0.7609 & 0.7643 & +0.0034 \\ 

        ToxCast & 0.6394 & 0.6525 & +0.0131 & 0.6395 & 0.6505 & +0.0110 \\ 
        \cline{1-7} 
        \specialrule{0em}{1pt}{0pt}
    \end{tabular} 
    }
    \label{tb_exp_pretrain_no_pretrain_gain_gcn_sage}
\end{table}

\subsection{Additional understanding and analysis for adaptive masking strategy} 
\label{sec_app_effectiveness_adam}
To evaluate the effectiveness of the proposed adaptive masking strategy, we compare the performance of GIN pre-trained by two different masking strategies (i.e., the uniform random masking strategy, denoted as \rdm\, and the adaptive masking strategy \am) on molecular classification datasets and summarize them in Table~\ref{tb_exp_pretrain_dy_mask_gain}. 
It can be seen that when using \am\ to select nodes to mask, the mask-and-predict paradigm may probably help the model learn more inherent graph-level molecular properties, thus getting better performance in downstream tasks (e.g., with maximum absolute ROC-AUC increase 11.4\% for the dataset ClinTox and 3.80\% in average over all of the molecular datasets). 

Therefore, we want to ask that what does \am\ bring to us, or what makes it different from the random masking strategy? 
One intuition is that if nodes that are less disturbed had larger probabilities to be masked in the current masking step, then nodes selected by \am\ will be distributed more evenly 
in the graph than nodes chosen by \rdm.
For further investigation, we calculate the average minimum distances between masked nodes\footnote{For each node $i$ in the masked nodes set $\mathcal{N}_s$, we calculate the minimum distance $d_{min}$ measured by the length of the shortest path between it and other nodes in the set (i.e., $d_{min} = \min \{ \text{dis} (i, j) \vert j \in \mathcal{N}_s, j \neq i \}$). Then the average value over all nodes in the masked nodes set is reported.} when adopting \rdm\ and \am\ with different masking steps ($k$) over the first five training epochs in Table~\ref{tb_exp_mask_node_dis}. All masking ratios are set to 15\%. Two observations can be made from Table~\ref{tb_exp_mask_node_dis}: 
1). The average minimum distances between masked nodes chosen by \am\ are larger than those of \rdm, which confirms our conjecture; 
2). The average minimum distance between masked nodes grows as the masking times $k$ increases. 
Hence, if we apply more masking steps, masked nodes will be more likely to distribute evenly over the molecule. 

\begin{table}[t]
    \centering
    \caption{\footnotesize The average minimum distances between nodes masked by the random masking strategy and our adaptive making strategy with different masking times $k$. Presented are the mean of the values calculated over the first five epochs.}
    \vspace{-2ex}
    \begin{tabular}{@{\;}c@{\;}|c|c|c|c|c@{\;}}
    \midrule
        \hline
        ~ & \rdm & $k = 3$ & $k = 5$ & $k = 6$ & $k = 7$ \\ \cline{1-6} 
        \specialrule{0em}{1pt}{0pt}
        
        Avg. Dis. & 2.7519 & 2.7584 & 2.7623 & 2.7625 & 2.7670 \\ 
        \cline{1-6} 
        \specialrule{0em}{1pt}{0pt}
    \end{tabular}
    \vspace{-2ex}
    \label{tb_exp_mask_node_dis}
\end{table}

\begin{table}[t]
    \centering
    \vspace{-2ex}
    \caption{Effectiveness of dynamic masking strategy v.s. uniformly random masking strategy. Bold numbers for those larger than $0.05$.}
    \begin{tabular}{@{\;}c@{\;}|c|c|c@{\;}}
    \midrule
        \hline
        ~ & \rdm & \am & \textbf{Abs. Imp.} \\ \cline{1-4} 
        \specialrule{0em}{1pt}{0pt}
        
        SIDER & 0.5947 & 0.6164 & +0.0217 \\ 
        
        ClinTox & 0.6685 & 0.7797 & \textbf{+0.1139} \\ 
      
        BACE & 0.8064 & 0.8224 & +0.0160 \\ 
        
        HIV & 0.7668 & 0.7704 & +0.0036 \\ 
        
        BBBP & 0.6316 & 0.7273 & \textbf{+0.0957} \\ %

        Tox21 & 0.7657 & 0.7696 & +0.0039 \\ 
        
        ToxCast & 0.6463 & 0.6603 & +0.0160 \\ 
        \cline{1-4} 
        \specialrule{0em}{1pt}{0pt}
    \end{tabular} 
    \vspace{-3ex}
    \label{tb_exp_pretrain_dy_mask_gain}
\end{table} 

\section{Further Analysis for Similarity-aware Sampling Strategy}

In this section, we conduct some further analysis of the proposed similarity-aware sampling strategy. 
This section is to support Sec.~\ref{sec_further_understanding_sim_aware_sampling}.

\subsection{Approximate similarity function}
\label{sec_app_assumption_sim_functions}
The pre-training graph dataset can be divided into two sets for each graph instance $G_i \in \mathcal{G}$ based on the ground-truth similarity function: 
$\text{sim}_{\text{gt}}(\cdot, \cdot)$ and a graph instance $G_i \in \mathcal{G}$:  $\mathcal{G}_i^{\text{gt}+} = \{ G_k | \text{sim}_{\text{gt}}(G_i, G_k) = 1, G_k \in \mathcal{G} \}$, containing $G_i$'s ground-truth positive instances and $\mathcal{G}_i^{\text{gt}-} = \{ G_k | \text{sim}_{\text{gt}}(G_i, G_k) = 0, G_k \in \mathcal{G} \}$, containing $G_i$'s negative graph instances. 

In practice, we use the approximate similarity function to approximate the ground-truth similarity function. 
We introduce the definition of such functions together with some reasonable properties assumed for their probability density functions over each $\mathcal{G}_i^{\text{gt}+}$ and $\mathcal{G}_i^{\text{gt}-}$ as follows: 
\begin{definition}[Approximation similarity function]
    For a similarity function $\text{sim}(\cdot, \cdot)$ and a graph instance $G_i \in \mathcal{G}$, denote its similarity score distribution density function over $G_i$'s ground-truth positive instance set $\mathcal{G}_i^{\text{gt}+}$ as $f_i^+(\cdot)$ and that over $\mathcal{G}_i^{\text{gt}-}$ as $f_i^-(\cdot)$. 
    Two properties are assumed for $f_i^+(\cdot)$ and $f_i^-(\cdot)$: 1). Both $f_i^+(\cdot)$ and $f_i^-(\cdot)$ are first-order differentiable functions over their domain of definition $[0, 1]$; 
    2). 
    There exists a similarity score threshold $0 < x_0 < 1$, s.t. for all $x_0 < x_1 < 1$, we have $\int_{x_1}^1 f_i^+(x) \mathrm{d}x > \int_{x_1}^1f_i^-(x) \mathrm{d}x$. 
    \label{def_approximation_similarity_function}
\end{definition}
Some specific probability density functions can be found for $f_i^+(\cdot)$ and $f_i^-(\cdot)$ such as normal distribution probability density functions (truncated between $[0, 1]$ and re-normalized by the integral over $[0, 1]$) and beta distribution probability density functions, in which cases the above mentioned properties can be easily satisfied by further restricting the relationship between their parameters. 

In our practice, another parameter -- a similarity threshold $\tau$ is introduced to divide the graph pre-training dataset $\mathcal{G}$ into two sets for each $G_i \in \mathcal{G}$: $\mathcal{G}_i^{\text{sim}_\tau+} = \{G_k | \text{sim}(G_i, G_k) \ge \tau, G_k \in \mathcal{G} \}$ and $\mathcal{G}_i^{\text{sim}_\tau-} = \{G_k | \text{sim}(G_i, G_k) < \tau, G_k \in \mathcal{G} \}$. 


\subsection{Why we should avoid false-positive instances in the contrastive learning process?} 
\label{sec_avoid_false_positives}
A graph instance $G_k$ is $G_i$'s false-positive instance if and only if $G_k\in \mathcal{G}_i^{\text{sim}_\tau +} \land G_k \notin \mathcal{G}_i^{\text{gt}+}$. 
It is intuitively correct that we should avoid such false-positive instances in the contrastive learning process. 
We investigate into such intuition by analyzing how false-positive instances would influence the contrastive training process in this section.

To get a glimpse into the training process, we start by introducing the loss function which is widely used in the contrastive learning process. 

The contrastive learning loss $\mathcal{L}$ is composed of losses from each graph instance in the pre-training dataset: 
\begin{equation}
    \mathcal{L} = \sum_{G_i\in \mathcal{G}} \mathcal{L}_i.
\end{equation}
where $G_i$ denotes a graph instance from the pre-training dataset. 
Moreover, we also use $x_j$ to denote the same instance (with $G_j$) in $G_i$'s positive sampling probability function $P_i^+(\cdot)$\footnote{To be more specific, $P_i^+(x_j)$ is the probability of sampling the graph instance $G_j$ as graph instance $G_i$'s positive instance in the contrastive learning process.} as well as the negative instance sampling probability function $P_i^-(\cdot)$ for simplicity. 
The actual value of $\mathcal{L}_i$ in each training epoch may be different from each other, since it is determined by the 
selected positive graph instance $G_i^+$ and each negative graph instance $G_i^-$. 
Thus, we use its expectation value here: 
\begin{equation}
    \mathbb{E}[\mathcal{L}_i] = -\mathbb{E}_{x_{i}^+\sim P_i^+} \log \left[ 
    \frac{\exp(z_i^T z_i^+ /\tau_t)}{\exp(z_i^T z_i^+ /\tau_t) + N \mathbb{E}_{x_{i}^- \sim P_i^-} \exp(z_i^T z_i^- /\tau_t)} 
    \right],
    \label{eq_regular_con_loss}
\end{equation}
where $\tau_t$ is the temperature hyper-parameter, $N$ is the number of negative instances sampled for each instance in one training epoch, $z_i = \frac{w_i}{\Vert w_i\Vert_2}\in \mathbb{R}^{K\times 1}$ is the normalized representation vector of graph instance $G_i$ with the feature dimension $K$, $w_i$ is the corresponding representation vector output by the neural network $\Vert w_i\Vert_2$ is the 2-norm of $w_i$. Here, we assume that negative instances are chosen from a uniform distribution over the whole pre-training dataset, which means that $P_i^-(x_j) = \frac{1}{\vert \mathcal{G} \vert}$ for each $x_i\in \mathcal{G}$ and $x_j \in \mathcal{G}$. 
Thus, two partial derivatives of interest are as follows: 
\begin{align}
    \mathbb{E}\left[\frac{\partial \mathcal{L}_j}{\partial w_i}\right] &= 
    -P_j^+(x_i) \frac{1}{\tau_t \Vert w_i\Vert_2}(1 - z_i z_i^T) \left[
    z_j - \frac{z_j \exp(z_j^T z_i /\tau_t) + \frac{N}{|\mathcal{G}|} z_j \exp(z_j^T z_i /\tau_t)}{\exp(z_j^T z_i /\tau_t) + 
    \frac{N}{\vert \mathcal{G} \vert} \sum_{x_j^- \in \mathcal{G}} 
    \exp(z_j^T z_j^- /\tau_t)} 
    \right]
    \\ 
    &= -P_j^+(x_i) \frac{1}{\tau_t \Vert w_i\Vert_2} (z_j - (z_i^T z_j)z_i) \left[
    \frac{\frac{N}{\vert \mathcal{G} \vert} \sum_{x_j^- \in \mathcal{G} \setminus \{ x_i \}} 
    \exp(z_j^T z_j^- /\tau_t)}{\exp(z_j^T z_i /\tau_t) +
    \frac{N}{\vert \mathcal{G} \vert}  \sum_{x_j^- \in \mathcal{G}} 
    \exp(z_j^T z_j^- /\tau_t)} 
    \right]
\end{align}
Let 
\begin{align}
    Q(x_j, x_i) &= \frac{\frac{N}{\vert \mathcal{G} \vert} \sum_{x_j^- \in \mathcal{G} \setminus \{ x_i \}} 
    \exp(z_j^T z_j^- /\tau_t)}{\exp(z_j^T z_i /\tau_t) + 
    \frac{N}{\vert \mathcal{G} \vert} \sum_{x_j^- \in \mathcal{G}} 
    \exp(z_j^T z_j^- /\tau_t)} \\ 
    P(x_j, x_i) &= \frac{\frac{N}{\vert \mathcal{G} \vert}
    \exp(z_j^T z_i /\tau_t)}{\exp(z_j^T z_i /\tau_t) + 
    \frac{N}{\vert \mathcal{G} \vert} \sum_{x_j^- \in \mathcal{G}} 
    \exp(z_j^T z_j^- /\tau_t)},
\end{align}
then we have: 
\begin{equation}
    \mathbb{E}\left[\frac{\partial \mathcal{L}_j}{\partial w_i}\right] = -P_j^+(x_i) \frac{1}{\tau_t \Vert w_i\Vert_2} (z_j - (z_i^Tz_j)z_i) Q(x_j, x_i)
    \label{eq_gradient_ji}
\end{equation}
As for $\mathbb{E}\left[\frac{\partial \mathcal{L}_i}{\partial w_i}\right]$, we have:
\begin{align}
    \mathbb{E}\left[ \frac{\partial \mathcal{L}_i}{\partial w_i} \right] &= -\frac{1}{\tau_t \Vert w_i \Vert_2} (1 - z_i z_i^T) \sum_{x_i^+ \in \mathcal{G}_i^+} P_i^+(x_i^+) \left(
    z_i^+ - \frac{z_i^+ \exp(z_i^T z_i^+ / \tau_t) + \frac{N}{\vert \mathcal{G} \vert} \sum_{x_i^-\in \mathcal{G}} z_i^- \exp(z_i^T z_i^- / \tau_t) }{\exp(z_i^T z_i^+ / \tau_t) + \frac{N}{\vert \mathcal{G} \vert } \sum_{x_i^- \in \mathcal{G}} \exp(z_i^T z_i^- / \tau_t)} 
    \right) 
    \label{eq_gradient_i_i_noapprox}
    \\ 
    &= -\frac{1}{\tau_t \Vert w_i \Vert_2} (1 - z_i z_i^T) \sum_{x_i^+\in \mathcal{G}_i^+} P_i^+(x_i^+) \left( z_i^+(P(x_i, x_i^+) + Q(x_i, x_i^+)) - 
    \sum_{x_i^-\in \mathcal{G}} z_i^- M^-(x_i, x_i^+, x_i^-) \right)
    \\ 
    &= -\frac{1}{\tau_t \Vert w_i \Vert_2} (1 - z_i z_i^T) \sum_{x_i^+\in \mathcal{G}_i^+} P_i^+(x_i^+) \left( z_i^+ (P(x_i, x_i^+) + Q(x_i, x_i^+)) \right) \\ 
    &+ \frac{1}{\tau_t \Vert w_i \Vert_2} (1 - z_i z_i^T) 
    \sum_{x_i^- \in \mathcal{G}} z_i^- \left( \sum_{x_i^+ \in \mathcal{G}_i^+} P_i^+(x_i^+) M^-(x_i, x_i^+, x_i^-) \right) 
    \\
    &= -\frac{1}{\tau_t \Vert w_i \Vert_2} (1 - z_i z_i^T) \sum_{x_i^+\in \mathcal{G}_i^+} P_i^+(x_i^+) \left( z_i^+ (P(x_i, x_i^+) + Q(x_i, x_i^+)) \right) \\
    &+ \frac{1}{\tau_t \Vert w_i \Vert_2} (1 - z_i z_i^T) 
    \sum_{x_i^- \in \mathcal{G}} z_i^- M(x_i, x_i^-),
    \label{eq_gradient_i_i_approx}
\end{align} 
where 
\begin{align}
    M^-(x_i, x_i^+, x_j) &= \frac{\frac{N}{\vert \mathcal{G} \vert} \exp(z_i^T z_j/ \tau_t)}{\exp(z_i^T z_i^+ / \tau_t) + \frac{N}{\vert \mathcal{G} \vert} \sum_{x_i^- \in \mathcal{G}} \exp(z_i^T z_i^- / \tau_t)},  
\end{align}
$M(x_i, x_j)$ is the weighted sum of $M^-(x_i, x_i^+, x_j)$, thus is the function of only $x_i$ and $x_j$: 
\begin{align}
    M(x_i, x_j) &= \sum_{x_i^+ \in \mathcal{G}_i^+} P_i^+(x_i^+) \frac{\frac{N}{\vert \mathcal{G} \vert} \exp(z_i^T z_j/ \tau_t)}{\exp(z_i^T z_i^+ / \tau_t) + \frac{N}{\vert \mathcal{G} \vert} \sum_{x_i^- \in \mathcal{G}} \exp(z_i^T z_i^- / \tau_t)} \\ 
    &= \sum_{x_i^+ \in \mathcal{G}_i^+} P_i^+(x_i^+) M^-(x_i, x_i^+, x_j).
\end{align}

Then, the expectation value of the partial gradient of the contrastive learning loss $\mathcal{L}$ on $G_i$'s representation vector $w_i$ brought by its positive instance $x_j$ is composed of the following two items: 
\begin{align}
    \mathbb{E}\left[ \frac{\partial \mathcal{L}_i}{\partial w_i} \right](x_j) &= -\frac{1}{\tau_t \Vert w_i \Vert_2}P_i^+(x_j) (z_j - (z_i^T z_j)z_i) (Q(x_i, x_j) + P(x_i, x_j)) \\ 
    \mathbb{E}\left[ \frac{\partial \mathcal{L}_j}{\partial w_i} \right] &= -\frac{1}{\tau_t \Vert w_i \Vert_2}P_j^+(x_i) (z_j - (z_i^T z_j)z_i) Q(x_j, x_i).
\end{align}

Following~\citep{khosla2020supervised}, we have: 
\begin{equation}
    \Vert z_j - (z_i^T z_j) z_i \Vert = \sqrt{1 - (z_i^T z_j)^2}, 
\end{equation}
which indicates that the magnitude of the partial gradient of contrastive learning loss on instance $x_i$'s representation vector $w_i$ 
from its positive instance $x_j$ is relevant with the cosine similarity ($z_i^T z_j$) between their representation vectors. 

Then, what we wish to explain here is that false-positive instances may have negative impact on the ability of the network to converge to the optimal state, which can be reached if we have the knowledge of each graph instance's ground-truth semantic class in the pre-training dataset. 

Let us assume that we have arrived at a training stage where the network has been optimized to a near-optimal state such that the cosine similarity between positive instance pairs' representation vectors are relatively high (i.e., $z_i^T z_i^+ \approx 1$), while those between negative instance pairs are relatively low (i.e., $z_i^T z_i^- \approx 0$). 
In this stage, the magnitude of the contrastive learning loss's gradient on instance $x_i$'s representation vector $w_i$ from its ground-truth positive instance $x_i^+$ may be relatively low given that $z_i^T z_i^+ \approx 1$. 
However, the magnitude of such gradient from its false-positive instance $x_j$ may be relatively high since $z_i^T z_j \approx 0$. 
Thus, the direction of $\frac{\partial \mathcal{L}}{\partial w_i}$ will be closer to that of $z_j - (z_i^T z_j)z_i$. 
It may be a bit deviated from the direction of the gradient from its ground-truth positive instance $z_i^+$ since: 
\begin{equation}
    (z_i^+ - (z_i^T z_i^+)z_i)^T (z_j - (z_i^T z_j)z_i) = z_i^{+T} z_j - z_i^{+T}z_i z_j^T z_i,
\end{equation}
which is near to zero since $z_i^T z_j \approx 0$ and $z_i^{+T} z_i \approx 1$. 

Such near orthogonal optimization direction 
implies the incorrect optimization direction and the unstable training process. 

\subsection{Higher similarity score may indicate higher probability to be a ground-truth positive instance} 
\label{sec_app_high_sim_high_gt_pos}
In this section, we want to investigate into whether it is reasonable to give graph instances that have higher similarity scores with the target graph instance $G_i$ larger probability to be sampled as its positive instances. 
We aim to find a similarity score interval such that graph instances that have larger similarity scores with $G_i$ means they also have higher probability to be $G_i$'s ground-truth positive instances when their similarity scores are changing in such an interval. 

Assume that we observe an instance $G_j$ whose similarity score with the target instance $G_i$ is $x$, denote the event that $G_j$ is a ground-truth positive instance of $G_i$ as $A$ and the event that $G_j$ is a negative instance of $G_i$ as $B$. Then, 
\begin{align}
    P(\text{sim} (G_i, G_j) = x | A) &= f_i^+(x) \delta x \\
    P(\text{sim} (G_i, G_j) = x | B) &= f_i^-(x) \delta x,
\end{align}
where $0 < \delta x \ll 1$ is a small quantity, $\text{sim}(\cdot, \cdot)$ is the similarity score function we use in practice, 
$f_i^+(\cdot)$ and $f_i^-(\cdot)$ are the corresponding similarity score probability density functions over $G_i$'s ground-truth positive instance set $\mathcal{G}_i^{\text{gt}+}$ and its negative instance set $\mathcal{G}_i^{\text{gt}-}$. 

Assume that we have no prior knowledge of the relationship between $G_i$ and $G_j$, which means that $P(A) = P(B) = \frac{1}{2}$.
Then, the posterior probabilities of the occurrence of the event $A$ and $B$ are as follows: 
\begin{align}
    P(A | \text{sim} (G_i, G_j) = x) &= \frac{f_i^+(x) \delta x}{f_i^+(x) \delta x + f_i^-(x) \delta x} = \frac{f_i^+(x)}{f_i^+(x) + f_i^-(x)} \\ 
    P(B | \text{sim} (G_i, G_j) = x) &= \frac{f_i^-(x) \delta x}{f_i^+(x) \delta x + f_i^-(x) \delta x} = \frac{f_i^-(x)}{f_i^+(x) + f_i^-(x)}.
\end{align}
The derivative of $P(A | \text{sim} (x_i, x_j) = x)$ with respect to the similarity score $x$ is: 
\begin{align}
    P(A | \text{sim} (x_i, x_j) = x)' = \frac{f_i^{+'}(x)f_i^-(x) - f_i^{-'}(x)f_i^+(x)}{(f_i^+(x) + f_i^-(x))^2}.
\end{align}
We wish to find the monotonic non-decreasing interval of $P(A|\text{sim}(x_i, x_j) = x)$ w.r.t. the similarity score $x$. 
The existence of such interval indicates that it is reasonable to give graphs instances that have higher similarity scores with the target instance larger probability to be selected as its positive instances. 

The question can be further changed to finding the similarity score interval where $\frac{f_i^{+'}(x)}{f_i^+(x)} > \frac{f_i^{-'}(x)}{f_i^-(x)}$ holds. 
Such interval may be determined by the shape of those two probability density functions and their respective parameters. 
Let us consider a specific function cluster: 
the truncated re-normalized normal distribution density function cluster. We explain such functions by giving an example as follows. Consider the probability density function of a normal distribution: $g_0(x) = \frac{1}{\sqrt{2\pi}\sigma}\exp\left(-\frac{(x-\mu)^2}{2\sigma^2}\right)$, its truncated re-normalized density function over $[0, 1]$ is $g(x) = \frac{g_0(x)}{\int_0^1 g_0(x) \mathrm{d}x}$. It is obvious that $\int_0^1 g(x) \mathrm{d}x = 1$. 
Consider the situation where both $f_i^+(\cdot)$ and $f_i^-(\cdot)$ are such truncated re-normalized normal distribution probability density functions with parameters $(\mu_+, \sigma_+)$ for $f_i^+(\cdot)$ and $(\mu_-, \sigma_-)$ for $f_i^-(\cdot)$.
It is naturally to assume that $\mu_- < \mu_+$ to meet with the properties proposed in Def.~\ref{def_approximation_similarity_function}.
Thus, we have: 
\begin{align}
    \frac{f_i^{+'}(x)}{f_i^+(x)} &= -\frac{x - \mu_+}{\sigma_+^2} \\ \frac{f_i^{-'}(x)}{f_i^-(x)} &= -\frac{x - \mu_-}{\sigma_-^2}.
\end{align}
Since we have no prior knowledge of the relationship between $\sigma_+$ and $\sigma_-$, we discuss the existence of the non-decreasing similarity score interval w.r.t. the relationship between $\sigma_+$ and $\sigma_-$ as follows: 
\begin{itemize}[noitemsep,topsep=0pt,parsep=0pt,partopsep=0pt,leftmargin=.5cm]
    \item Case 1. If $\sigma_+ = \sigma_-$, we have $\frac{f_i^{+'}(x)}{f_i^+(x)} > \frac{f_i^{-'}(x)}{f_i^-(x)}$ for every $0 \le x \le 1$; 
    \item Case 2. If $\sigma_+ < \sigma_-$, which indicates that the similarity score distribution over the ground-truth graph instance set is more centralized, $\frac{f_i^{+'}(x)}{f_i^+(x)} > \frac{f_i^{-'}(x)}{f_i^-(x)}$ can be satisfied when $0\le x < \min\left( \frac{\sigma_-^2 \mu_+ - \sigma_+^2\mu_-}{\sigma_-^2 - \sigma_+^2}, 1 \right)$; 
    \item Case 3. If $\sigma_+ > \sigma_-$, $\frac{f_i^{+'}(x)}{f_i^+(x)} > \frac{f_i^{-'}(x)}{f_i^-(x)}$ can be satisfied when $\max\left(\frac{\sigma_+^2\mu_- - \sigma_-^2\mu_+}{\sigma_+^2 - \sigma_-^2}, 0 \right) < x \le 1$. 
\end{itemize}

The limitation here is that we restrict the shape of the similarity score possibility distribution density functions as well as the relationship between their parameters to arrive at the above conclusion. 
It is possible that similar conclusions can be arrived at when $f_i^+(\cdot)$ and $f_i^-(\cdot)$ are in other forms. 


\subsection{Sampling preference brought by the high-order graph sampling strategy}
\label{sec_preference_of_ho_sampling}
In this section, we want to discuss into the sampling preference brought by the high-order sampling process towards those graph instances that are high-order connected to the target graph instance, including 
those graph instances that are both high-order connected and also lower-order connected to the target graph instance as well as those that are only high-order connected to the target graph instance. 

The connectivity between two nodes in the graph is introduced as follows: 
\begin{definition}[Connectivity]
    Node $n_i$ and node $n_j$ is $k$-connected, if and only if there exists a loop-free path with length $k$ between them. For such two nodes $n_i$ and $n_j$, a node sequence $n_0 (=n_i), n_1, ..., n_k (=n_j)$ can be found, where $n_{p} \neq n_{q}, \forall p \neq q, 0\le p\le k, 0\le q\le k$. $k$ is a connectivity order between node $n_i$ and node $n_j$. 
\end{definition}

We only take the second-order sampling strategy as an example here for its simplicity and generality. It is easy to generalize to high-order sampling strategy, since the properties of the high-order random walk, the high-order sampling strategy used in this paper, have been 
researched and discussed thoroughly. 
The ratio for selecting each first-order neighbour $G_j \in \mathcal{G}_i^{\text{sim}_\tau+}$ of the graph instance $G_i$ when using the first-order sampling strategy is:
\begin{equation}
    P_i^+(x_j) = \frac{\text{sim}(G_i, G_j)}{\sum_{G_k \in \mathcal{G}_i^{\text{sim}_\tau+} \text{sim}(G_i, G_k) } }.
\end{equation}
For each graph instance $G_i^-\in \mathcal{G}\setminus \mathcal{G}_i^{\text{sim}_\tau+}$, we have $P_i^+(x_i^-) = 0$.

When using second-order sampling strategy, the corresponding sampling ratio is proportional to: 
\begin{equation}
    \hat{P}_i^{2+}(x_j) = P_i^+(x_j) + \sum_{G_k\in \mathcal{G}_i^{\text{sim}+}} P_i^+(x_k) P_k^+(x_j), 
\end{equation}
for each $G_j \in \mathcal{G}$.

Obviously, second-order sampling strategy gives larger preference for $G_i$'s first-order neighbours that are also 2-connected to $G_i$ than its neighbours that are only 1-connected to $G_i$, compared with the first-order sampling strategy\footnote{Note that this does not mean that the sampling ratio for neighbours that are both 1-connected and 2-connected to $G_i$ is larger than that for neighbours that are only 1-connected to $G_i$.}. 

\vpara{Experimental evidence.}
Such sampling preference to 2-connected first-order neighbours can be verified by simple experimental results from the following aspects: 1). Similarity scores between $G_i$'s first-order neighbours that are connected to each other selected by the second-order sampling strategy should be higher than that resulted by the first-order sampling strategy. It is because that if second-order sampling strategy tends to select 2-connected first-order neighbours more than only 1-connected neighbours, it is more likely that the chosen first-order neighbours are also connected with each other when using the second-order sampling strategy. 
It can be verified by experimental results shown in 
Table~\ref{tb_exp_sta_first_second_sampling_sim_con_ratio}, the sampled first-order neighbours are more likely to be connected to each other and also more similar with each other when using second-order sampling strategy than using first-order sampling strategy. 
2). Sampled first-order neighbours tend to be more similar with the target graph instance when using second-order sampling strategy than using first-order sampling strategy. 
It is not a straightforward conclusion that can be drawn by analyzing the sampling preference of the second-order sampling strategy for different kinds of neighbours, but can be seen from the experimental results (see the column ``Target Sim.'' in Table~\ref{tb_exp_sta_first_second_sampling_sim_con_ratio}). 
Thus, the second-order sampling strategy may also tend to sample more similar first-order neighbours, which may be more likely to be ground-truth positive instances. 

\begin{table}[t]
    \centering
    \vspace{-2ex}
    \caption{\footnotesize Statistical results for similarity scores within sampled positive instances, similarity scores for sampled positive instances with the target graph instance and the ratio of sampled positive instances connected to each other. Results for first-order neighbourhood sampling strategy and second-order sampling strategy are presented in the table, where only sampled first-order neighbours are chosen for calculation. In the experiment, we uniformly randomly sample $1000$ molecules from the pre-training dataset and calculate the mean value of those three values. Three independent experiments were performed with mean and standard deviation values reported. Values presented in the table have the format $\text{mean}\pm \text{std}$. For abbreviations used, ``Inter-pos. Sim.'' denotes the similarity scores within sampled positive instances, ``Target Sim.'' denotes the similarity scores for sampled instances with the target graph instance.}
    \begin{tabular}{@{\;}c@{\;}|c|c|c@{\;}}
    \midrule
        \hline
        ~ & Inter-pos. Sim. & Target Sim. & Connected Ratio  \\ \cline{1-4} 
        \specialrule{0em}{1pt}{0pt}
        
        First-order & $0.4552\pm 0.0026$ & $0.4766\pm 0.0035$ & $0.6171\pm 0.0108$ \\ 
        
        Second-order & $0.4687\pm 0.0057$ & $0.4831\pm 0.0022$ & $0.6857\pm 0.0302$ \\ 
        \cline{1-4} 
        \specialrule{0em}{1pt}{0pt}
    \end{tabular} 
    \vspace{-3ex}
    \label{tb_exp_sta_first_second_sampling_sim_con_ratio}
\end{table} 

However, second-order sampling also leads to the possibility to sample neighbours that are only 2-connected to the target graph instance, whose sampling rates are proportional to: 
\begin{equation}
    \hat{P}_i^{2+}(x_j) = \sum_{G_k\in \mathcal{G}_i^{\text{sim}+}} P_i^+(x_k) P_k^+(x_j). 
\end{equation}
Since such instances that are only 2-connected to the target graph instances  
are more likely to be false-positive instances, high-order sampling strategy may increase the risk of sampling false-positive instances. 

\subsection{Sampling bias brought by the approximate similarity score function}
\label{sec_balance_pos_neg}

In this section, we want to discuss the possible sampling bias brought by the approximate similarity score function. 
The sampling bias may exist in two aspects: 
1). Graph instance $G_i$'s ground-truth positive instances that have higher similarity scores with $G_i$ will enjoy larger sampling preference.
Moreover, $G_i$'s ground-truth positive instances with relatively low similarity scores are failed to be selected as its positive instances. 
However, they should be sampled equally when using the ground-truth similarity score function. 
2). 
It is possible that $G_i$'s negative instances that have relatively high similarity scores could be selected as its positive instances.

If we denote $P_i^{\text{gt}+}(\cdot)$ as graph $G_i$'s ground-truth positive sampling ratio function, then $G_i$'s positive sampling bias when using the approximate similarity function from the ground-truth similarity function is defined as:
\begin{align}
    \text{bias}_i^\tau &= \sum_{G_k\in \mathcal{G}} \vert P_i^+(x_k) - P_i^{\text{gt}+}(x_k) \vert \\ 
    &= \sum_{G_k\in \mathcal{G}_i^{\text{gt}+}} \vert P_i^+(x_k) - P_i^{\text{gt}+}(x_k) \vert + \sum_{G_k\in \mathcal{G}_i^{\text{gt}-}} \vert P_i^+(x_k) - P_i^{\text{gt}+}(x_k) \vert \\ 
    &= \text{gap}_i^\tau + \text{risk}_i^\tau.
\end{align}
$\text{gap}_i^\tau$ and $\text{risk}_i^\tau$ are the functions of $f_i^+(\cdot)$, $f_i^-(\cdot)$ and $\tau$:
\begin{align}
    \text{gap}_i^\tau &= \vert \mathcal{G}_i^{\text{gt}+}\vert \int_0^\tau \frac{f_i^+(x)}{\vert \mathcal{G}_i^{\text{gt}+} \vert} \mathrm{d}x + \vert \mathcal{G}_i^{\text{gt}+} \vert  \int_\tau^1 \left\vert \frac{x }{\text{totsim}_\tau} - \frac{1}{\vert \mathcal{G}_i^{\text{gt}+} \vert}  \right\vert f_i^+(x) \mathrm{d}x \\ 
    \text{risk}_i^\tau &= \vert \mathcal{G}_i^{\text{gt}-} \vert  \frac{ \int_\tau^1 xf_i^-(x) \mathrm{d}x}{\text{totsim}_\tau},
\end{align}
where $\text{totsim}_\tau = |\mathcal{G}_i^{\text{gt}+}| \int_\tau^1 x f_i^+(x)\mathrm{d}x + |\mathcal{G}_i^{\text{gt}-}| \int_\tau^1 xf_i^-(x)\mathrm{d}x$ is the sum of the similarity scores over $G_i$'s positive instance candidates. 
Assume that: 
\begin{itemize}[noitemsep,topsep=0pt,parsep=0pt,partopsep=0pt,leftmargin=.5cm]
    \item Ground-truth positive instances would always not be explored thoroughly, which means that $P_i^+(x_k) = \frac{\text{sim}(x_i, x_k)}{\text{totsim}_\tau} > \frac{1}{\vert \mathcal{G}_i^{\text{gt}+}\vert }$ for each $G_k \in \mathcal{G}_i^{\text{gt}+} \cap \mathcal{G}_i^{\text{sim}_\tau +}$. 
\end{itemize}
This assumption is reasonable, since
\begin{itemize}[noitemsep,topsep=0pt,parsep=0pt,partopsep=0pt,leftmargin=.5cm]
    \item $\vert \mathcal{G}_i^{\text{gt}+} \vert \gg \vert \mathcal{G}_i^{\text{sim}_\tau+} \vert$, considering that the pre-training dataset is always large and while $\vert \mathcal{G}_i^{\text{sim}_\tau+} \vert$ is relatively small to reduce the similarity score computation budget for efficiency. 
\end{itemize}
To meet with the assumption, we can further introduce $\tau_3$ for each graph instance $G_i$, where $\frac{\tau_3}{
\text{totsim}_{\tau_3}
} = \frac{1}{\vert \mathcal{G}_i^{\text{gt}+} \vert}$, and restrict the similarity threshold $\tau$ to $\tau_3 < \tau < 1$ 
for each $G_i$'s $\tau_3$. 

Thus, $\text{gap}_i^\tau$ becomes:
\begin{align}
    \text{gap}_i^\tau &= \vert \mathcal{G}_i^{\text{gt}+}\vert  \int_0^\tau \frac{f_i^+(x)}{\vert \mathcal{G}_i^{\text{gt}+} \vert} \mathrm{d}x + \vert \mathcal{G}_i^{\text{gt}+} \vert  \int_\tau^1 \left( \frac{x}{\text{totsim}_\tau} - \frac{1}{\vert \mathcal{G}_i^{\text{gt}+} \vert}  \right) f_i^+(x) \mathrm{d}x \\ 
    &= \left \{ \int_0^\tau f_i^+(x)\mathrm{d}x - \int_\tau^1 f_i^+(x)\mathrm{d}x \right \} 
    + \left \{ \vert \mathcal{G}_i^{\text{gt}+} \vert  \int_\tau^1 \frac{x f_i^+(x)}{\text{totsim}_\tau} \mathrm{d}x \right \} 
    \label{eq_expansion_last_gap_tau}
\end{align}
We denote such two items as $\text{gap}_i^{\tau,\text{gt}} = \int_0^\tau f_i^+(x)\mathrm{d}x - \int_\tau^1 f_i^+(x)\mathrm{d}x $ and $\text{gap}_i^{\tau,\text{sim}_\tau} = \vert \mathcal{G}_i^{\text{gt}+}\vert \cdot \int_\tau^1 \frac{x f_i^+(x)}{\text{totsim}_\tau}\mathrm{d}x$ respectively. 

If we assume that $f_i^+(\cdot)$ and $f_i^-(\cdot)$ are truncated re-normalzied normal distribution functions and further restrict the relationship between their parameters, we can show that $\text{risk}_i^\tau$ increases as the hreshold $\tau$ increases, thus at the same time $\int_\tau^1\frac{x f_i^+(x)}{\text{totsim}_\tau}\mathrm{d}x$ decreases as $\tau$ increases since $\frac{\vert \mathcal{G}_i^{\text{gt}+} \vert \int_\tau^1 xf_i^+(x) \mathrm{d}x}{\text{totsim}_\tau} + \frac{\vert \mathcal{G}_i^{\text{gt}-} \vert \int_\tau^1 xf_i^-(x) \mathrm{d}x}{\text{totsim}_\tau} = 1$, though may not that intuitive: 
\begin{align}
    \text{risk}_i^\tau &= \frac{\vert \mathcal{G}_i^{\text{gt}-} \vert \int_\tau^1 xf_i^-(x) \mathrm{d}x}{\vert \mathcal{G}_i^{\text{gt}-} \vert \int_\tau^1 xf_i^-(x) \mathrm{d}x + \vert \mathcal{G}_i^{\text{gt}+} \vert \int_\tau^1 xf_i^+(x) \mathrm{d}x} \\ 
    \frac{\partial \text{risk}_i^\tau}{\partial \tau} &= 
    \frac{\vert \mathcal{G}_i^{\text{gt}+} \vert \vert \mathcal{G}_i^{\text{gt}-} \vert \tau (f_i^+(\tau) \int_\tau^1 x f_i^-(x)\mathrm{d}x - f_i^-(\tau) \int_\tau^1 x f_i^+(x)\mathrm{d}x )}{(\vert \mathcal{G}_i^{\text{gt}-} \vert \int_\tau^1 xf_i^-(x) \mathrm{d}x + \vert \mathcal{G}_i^{\text{gt}+} \vert \int_\tau^1 xf_i^+(x) \mathrm{d}x)^2},
\end{align}
where the sign of $\frac{\partial \text{risk}_i^\tau}{\partial \tau}$ is determined by the sign of $f_i^+(\tau) \int_\tau^1 x f_i^-(x)\mathrm{d}x - f_i^-(\tau) \int_\tau^1 x f_i^+(x)\mathrm{d}x$. 
For $f_i^+(\cdot)$'s parameters $(\mu_+, \sigma_+)$ and $f_i^-(\cdot)$'s 
parameters $(\mu_-, \sigma_-)$, we assume that $\sigma_+ = \sigma_- = \sigma, \mu_+ > \mu_-$. 
Thus, 
\begin{align}
    \frac{\int_\tau^1 xf_i^-(x) \mathrm{d}x}{f_i^-(\tau)} &= \frac{\int_\tau^1 x\exp\left(\frac{(x - \mu_-)^2}{2\sigma^2}\right) \mathrm{d}x}{\exp\left(\frac{(\tau - \mu_-)^2}{2\sigma^2}\right)} 
    = \int_\tau^1 x\exp\left( \frac{x^2 - 2\mu_-(x - \tau) - \tau^2}{2\sigma^2} \right) \mathrm{d} x \\ 
    \frac{\int_\tau^1 x f_i^+(x) \mathrm{d}x}{f_i^+(\tau)} &= \frac{\int_\tau^1 x\exp\left(\frac{(x - \mu_+)^2}{2\sigma^2}\right) \mathrm{d}x}{\exp\left(\frac{(\tau - \mu_+)^2}{2\sigma^2}\right)}  
    = \int_\tau^1 x\exp\left( \frac{x^2 - 2\mu_+(x - \tau) - \tau^2}{2\sigma^2} \right) \mathrm{d} x. 
\end{align}
We have, 
\begin{equation}
    \int_\tau^1 x\exp\left( \frac{x^2 - 2\mu_-(x - \tau) - \tau^2}{2\sigma^2} \right) \mathrm{d} x > \int_\tau^1 x\exp\left( \frac{x^2 - 2\mu_+(x - \tau) - \tau^2}{2\sigma^2} \right) \mathrm{d} x 
\end{equation} 
since $x - \tau \ge 0, \forall \tau\le x\le 1$ and $\mu_- < \mu_+$. 

Thus, $\text{gap}_i^{\tau,\text{gt}}$ and $\text{risk}_i^\tau$ decrease as $\tau$ decreases, while $\text{gap}_i^{\tau,\text{sim}_\tau}$ increases as $\tau$ decreases. 
Though it is hard to detect the monotonicity of $\text{gap}_i^\tau$, we can get a sense that there exists a balance between minimizing $\text{gap}_\tau$ and $\text{risk}_\tau$. 

A limitation that must be pointed out about the above conclusion,
including the monotonicity of each part in $\text{gap}_i^\tau$ and $\text{risk}_i^\tau$, is arrived by assuming the similarity possibility density functions $f_i^+(\cdot)$ and $f_i^-(\cdot)$ are chosen from a certain function cluster and further restricting the relationship between their parameters. 
Thus, the above conclusion only aims at giving a glimpse into the potential balance existing in the sampling bias brought by the approximate similarity function over the ground-truth positive instances and negative instances. 

Though it seems that the high-order sampling strategy is not taken into consideration in the above discussion, it can be naturally integrated in since the positive similarity threshold $\tau$ is lowered in the high-order sampling process. 
Moreover, the abstracted similarity threshold $\tau$ implies that we can probably reach a good balance point, which may be determined by the specific application scenario, by tuning the similarity score threshold $\tau$ directly or changing the positive instance sampling strategy.

\section{Broader Impact}
\label{sec_broader_impact}
In this paper, we have developed a sampling based graph positive instance selection strategy (\hgcn) that can be used in the graph contrastive learning process. Compared with previous approaches, where positive instances for the target graph instance are obtained by performing data augmentation skills on the target graph instance, our sampling based process can ultimately get positive graph instances of better quality keeping enough similarity with the target graph instance and also within different positive graph instances. Such a sampling process can also ensure the necessary domain specific information preserved in the resulting positive instances. We also propose an improvement on a widely used node-level pre-training strategy (\AM), leading to masked nodes distributed more evently on the graph. Moreover, we discover the potential possibility of the pre-trained GNN models to perform cross-domain transferring. 

It seems that there are no explicit relationship between our graph-level similarity based positive instance sampling strategy and our improvement for the widely used attribute masking pre-training strategy.
However, we want to point out that their design principles point towards a higher methodology design philosophy. That is, introducing prior knowledge into methods that are originally random ones can help us get better results. Specifically, we introduce the approximate pair-wise similarity information which can help us sample positive instances of better quality. By comparison, previous methods always tend to use data augmentation methods to construct positive graph instances from the target instance. Such strategies always introduce random factors to perturb the structure and attribute information in the graph. Those random factors may destruct necessary information that should be kept in the positive instances. Admittedly, it is also possible to design data augmentation strategies that are aware of such information to preserve them in the resulting graph instances. However, it will turn out to be a complex strategy with many restrictions on the augmentation process. The advantage of our \hgcn\ is then obvious -- it keeps such necessary information in the positive samples by sampling from existing graphs with some deterministic factors fused into the sampling process automatically (transition probability from one node in the graph to another node is calculated by their pair-wise similarity score). Thus, it is a more effective, efficient and elegant solution for such a crucial problem existing in contrastive learning for graph data. 

Our adaptive masking strategy tries to select nodes based on their perturbation scores, which ultimately leads to nodes selected distributed more evenly in the graph. It is inspired by Kmeans++~\citep{arthur2006k}. Though we still aim at distributing nodes evenly in the graph, we approximately solve such a problem by adding some deterministic factors (the sampling rates for remaining nodes are calculated based on their perturbation scores) in the node selecting process, different from previous methods which just selecting nodes according to a uniform distribution. 

The broader impact of our research can be summarized below: 
\begin{itemize}[noitemsep,topsep=0pt,parsep=0pt,partopsep=0pt,leftmargin=.5cm]
    \item \textbf{For machine learning community:} This work demonstrates the importance of designing machine learning strategies by thinking deeply into essential things that are most important to solve the problem (e.g., how to ensure the enough similarity between positive instances and target instances in our positive instance selection problem). The sampling based positive instance selection process may probably inspire more novel graph instance pre-training strategies. 
    
    Moreover, we point out a potential new developing direction for pre-training on graph data. That is, how can we obtain powerful universally transferrable pre-trained GNN models that can transfer across different kinds of graphs? It is an interesting and also a valuable quation that deserves further discussion. 
    \item \textbf{For the drug discovery community:} Researchers from the drug discovery community can benefit from this work. It is because that the starting point of the design of \hgc\ is the wish to apply contrastive learning strategy on molecular graphs effectively, since previous approaches may impede the development of contrastive learning for pre-training on molecular graphs, which are kind of special graphs in the real-world. Thus, the contrastive learning using \hgc\ for positive instance sampling can help with developing GNN pre-training strategies. We hope that \hgcn\ can help with boosting the performance of various drug discovery applications, such as molecular property prediction and virtual screening.
\end{itemize}

\end{document}